\pdfoutput=1

\documentclass[11pt]{article}

\usepackage{EACL2023}

\usepackage{times}
\usepackage{latexsym}

\usepackage[T1]{fontenc}

\usepackage[utf8]{inputenc}

\usepackage{microtype}
\usepackage{booktabs} 
\usepackage{graphicx}
\usepackage{wrapfig}
\usepackage{placeins}

\newcommand{\linebreakcell}[1]{
  \begin{tabular}{c}#1\end{tabular}}
\newcommand{\revised}[1]{\textcolor{black}{#1}}

\usepackage{inconsolata}

\title{Know your audience: specializing grounded language models with listener subtraction}

\author{
Aaditya K Singh \\ Gatsby Computational Neuroscience Unit \\ University College London \\ London, W1T 4JG \\ \texttt{aaditya.singh.21@ucl.ac.uk}
\And
David Ding \\ DeepMind \\ London, UK \\ \texttt{fding@deepmind.com}
\AND
Andrew Saxe \\ Gatsby Computational Neuroscience Unit \\ University College London \\ London, W1T 4JG \\ \texttt{a.saxe@ucl.ac.uk}
\And
Felix Hill \\ DeepMind \\ London, UK \\ \texttt{felixhill@deepmind.com}
\AND
Andrew Kyle Lampinen \\ DeepMind \\ London, UK \\ \texttt{lampinen@deepmind.com}
}

\begin{document}
\maketitle

\begin{abstract}
Effective communication requires adapting to the idiosyncrasies of each communicative context---such as the common ground shared with each partner. Humans demonstrate this ability to specialize to their audience in many contexts, such as the popular game Dixit.
We take inspiration from Dixit to formulate a multi-agent image reference game where a (trained) speaker model is rewarded for describing a target image such that one (pretrained) listener model can correctly identify it among distractors, but another listener cannot.
To adapt, the speaker must exploit differences in the knowledge it shares with the different listeners. 
We show that finetuning an attention-based adapter between a CLIP vision encoder and a large language model in this contrastive, multi-agent setting gives rise to context-dependent \emph{natural language} specialization from rewards only, without direct supervision. 
Through controlled experiments, we show that training a speaker with two listeners that perceive differently, using our method, allows the speaker to adapt to the idiosyncracies of the listeners. Furthermore, we show zero-shot transfer of the specialization to real-world data. 
Our experiments demonstrate a method for specializing grounded language models without direct supervision and highlight the interesting research challenges posed by complex multi-agent communication. 
\end{abstract}

\section{Introduction}

Human language use is communicative, and thus involves substantial adaptation to each conversational partner and context \citep{clark1986referring,clark1996using,communicationAccommodationTheory,frank2012predicting, hawkins2019emergence, hawkins2022partners}. We can adapt our speech to complex social settings, with multiple partners \citep{mankewitz2021,boyce2022} and competing constraints, such as politeness vs. explicitness \citep{yoon2016talking}. While current language and captioning models can increasingly imitate human language and scene descriptions \citep{gpt3, frozenFSL, clipcap, flamingo, imageCaptioningReview}, they generally are not explicitly adapted to a particular partner or to satisfy multiple constraints. Common approaches to specializing the language generated by these models fall into two categories: finetuning on a supervised dataset \citep[e.g.][]{lora}, which requires task-specific labeled examples, or prompt engineering, which is often brittle and may still require task-specific examples \citep{liu2021pre}. Here, we take inspiration from how humans adapt their language to their audience to offer an alternative approach to specializing grounded language models without direct supervision.

\begin{figure*}[t]
    \centering
    \includegraphics{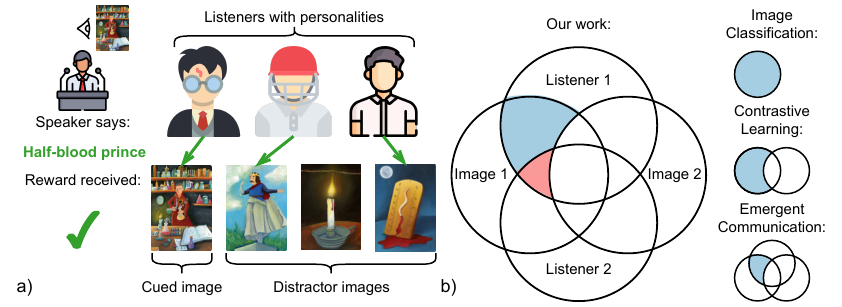}
    \vspace{-0.75em}
    \caption{a) An example round of Dixit, the inspiration for our work. The speaker uses a specialized caption (green), so that only one listener (who has read Harry Potter) correctly identifies the image. This outcome is rewarded per the \textit{some but not all} rule. b) How our work relates to past work.
    Image classification: models say something about an image. Contrastive learning: models say something to distinguish an image from others. Emergent communication: models say something to distinguish an image from others in a way one listener will understand. Our work: models say something to distinguish an image from others in a way one \textit{but not another} listener will understand.}
    \label{fig:motivation}
    \vspace{-1.2em}
\end{figure*}

Our approach to specialization is inspired by the popular multi-player game Dixit \citep{dixitgame}, which rewards each player's ability to adapt their communication to a particular audience. In a round of Dixit, the ``speaker'' describes a chosen image in language. Then, the rest of the players (two or more ``listeners'') attempt to identify the target image from a pool of distractors. Importantly, the speaker is rewarded when \textit{some but not all} listeners correctly identify the image. This leads to creative captions that target the common ground between a speaker and some (but not all) of the listeners (see example in Figure \ref{fig:motivation}a). For more details, we refer readers to \citet{dixitchallenge}, who introduce Dixit as a grand challenge for AI.

Using this inspiration, we formulate a setting where grounded language models can specialize their language without direct supervision, from rewards only. We build on prior work in emergent communication \citep{dwd}, and adapt a pretrained captioning model (the ``speaker'') by finetuning a small fraction of its parameters to maximize a Dixit-like reward---this minimal adaptation helps reduce language drift \citep{counteringdrift}. We go beyond prior work by investigating a complex, not fully cooperative, setting (Figure \ref{fig:motivation}b). The speaker's goal is to communicate information to some listeners, but not others, by exploiting differences in listeners' idiosyncratic ``personalities''. We instantiate this setting by having a speaker model communicate with multiple (frozen) contrastive listener models differing along perceptual axes. We design careful experiments, with datasets, metrics, and controls to ensure quantifiable assessment of language specialization.

We show that training a speaker with a pair of listeners in this Dixit-inspired setting can lead to \textit{diverse} specialization of natural language, with minimal language drift, and without direct supervision.
From rewards, the speaker identifies and learns to cue to the difference between two listeners,
which we call ``listener subtraction''. We show this adaptation across many pairs of perceptually-differing listeners, and various datasets (Section \ref{sec:fmnist}, \ref{sec:coco}). For example, when trained with one listener which sees color, and another which sees only grayscale, the speaker learns to exclusively use color words---it exploits the specialized information it shares with the first listener. Furthermore, the speaker exhibits some zero-shot transfer of its language specialization from artificial to realistic datasets (Section \ref{sec:zeroshot}). To our knowledge, our work is the first to consider natural language communication in a grounded, multi-listener setting, without direct supervision.

\section{Related work}
\textbf{Dixit}. \citet{dixitchallenge} discuss the full game of Dixit as a grand challenge for AI and posit the various interesting subproblems that would need to be solved (but do not attempt to solve them). Here, we focus on addressing the ``Find a Phrase'' subproblem---how the speaker should choose a phrase such that some but not all listeners correctly identify the image. 

\textbf{Adapting (grounded) language models}. A large body of prior work focuses on adapting large pretrained language models \citep{lora, ziegler2020, gpt3, frozenFSL} to various tasks. Our work also focuses on natural language specialization, but differs in that we use no supervised data. In that sense, our work is more similar to approaches that fine-tune pretrained models with reinforcement learning \citep{stiennon2020learning,ouyang2022training}. Unlike those works, we focus on grounded communication. With respect to grounding, we take inspiration from prefix tuning \citep{prefixtuning} and use an image-to-prefix encoder to condition a pretrained language model on images, as in \citet{frozenFSL}. These models can adapt their language via prompting  \citep{gpt3}, which we consider as a baseline for our approach.

\textbf{Pragmatics through model interactions}. Work in cognitive science has explored how pragmatic inferences can be explained as reasoning over simpler speaker and listener models \citep{frank2012predicting}. This perspective aims to explain human pragmatic references in grounded reference games and can help train models that give and follow instructions \citep{fried-etal-2018-unified} or \revised{help train listener models that adapt to individual human speakers \citep{hawkins-etal-2020-continual}. Our work is related to this framework, but differs in that the \textit{speaker model} is adapted to use \textit{differences between multiple listeners}.}

\textbf{Emergent communication}. Finally, there is prior work on emergent communication in image reference games. We focus on works involving natural language generation, like our setting. \citet{dwd} focus on a speaker agent that asks questions in dialogue with a single, pretrained-and-frozen listener, to identify an image among distractors. \citet{angeliki2020} consider speaker and listener agents in a more standard image reference game---the speaker describes an image, and the single listener must choose the image among distractors. Both papers use the pretrain-then-finetune approach to minimize language drift; we adopt a similar methodology. Like \citet{dwd}, we freeze our listeners to maintain language grounding \citep[cf.][]{counteringdrift}, and focus on the speaker. Unlike prior work, our setting involves simultaneous, natural language communication to multiple listeners and is not fully cooperative.

\section{Methods}
\label{sec:methods}
\subsection{Multi-agent image reference game} \label{sec:multiagentgame}
We focus on a multi-agent image reference game inspired by a single round of Dixit. We frame this problem as a challenge of tuning a pretrained multimodal model to specialize its language, without direct supervision on how to specialize. Our setup (Figure \ref{fig:arch}) involves one speaker model communicating with multiple listener models about a target image. The speaker model receives the target image and outputs a caption. Each listener model receives the caption and the target image along with some (random) distractor images. Each listener model's goal is to correctly identify the target image from the distractors, based on the speaker's caption.

Following the some-but-not-all reward structure of the Dixit game, we reward the speaker model for \textit{the difference in the (binary) accuracies of listeners 1 and 2 on a given set of images}. To avoid co-adaptation and pragmatic drift \citep[cf.][]{angeliki2020}, we pretrain and freeze our listener models. Thus, only the speaker has learnable parameters.

\subsection{Speaker model}
\label{sec:speaker}
\begin{figure}[t]
  \begin{minipage}[t]{\linewidth}
    \includegraphics[width=\linewidth]{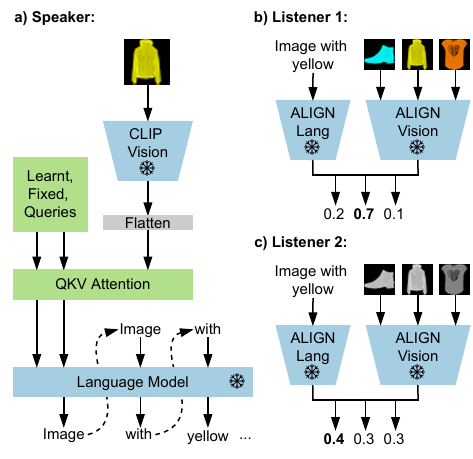}
    \vskip-0.75em
    \caption{Our setup for cFMNIST and the grayscale transform. a) The speaker receives an image and generates a caption. b) Listener 1 receives unperturbed images and picks best match to the caption (correctly). c) Listener 2 receives images with the grayscale transform applied and picks the best match to the caption (incorrectly). Then, the speaker receives a reward of 1 (the difference in the accuracies) which it uses to update its adapter parameters (green) via REINFORCE. All components in blue are pretrained and frozen.}
    \label{fig:arch}
  \end{minipage}
  \vskip-1.5em
\end{figure}

The speaker model (Figure \ref{fig:arch}a) is inspired by the \textit{Frozen} model \citep{frozenFSL} and ClipCap \citep{clipcap}. We build the speaker using several pretrained components, and adapt only a small piece to our setting.
The model is composed of a pretrained-and-frozen CLIP visual encoder \citep{clip}, learned attentive (QKV) adapter layer, and a pretrained-and-frozen Transformer (decoder) language model \citep{transformer, chinchilla}. An input image is first passed through the visual encoder. The unpooled output of the encoder is flattened and passed into the adapter, a Perceiver-inspired cross attention layer \citep{percieverio}. The adapter outputs $n=32$ prefix tokens, which are used to condition the language model and generate up to 32 output tokens. See Appendix \ref{appx:arch-details} for more details.

\emph{The only parameters of this model that are ever trained are in the adapter.}
We keep the pretrained CLIP encoder and language model frozen (to help prevent language drift), and pretrain the adapter on the Conceptual Captions dataset \citep{conceptualCaptions}
to give our speaker basic captioning abilities (see Appendix \ref{appx:speaker-pretrain}). All our experiments start from this captioning-pretrained base speaker model.
We finetune the adapter for each experimental condition, which consists of a dataset to train on and a pair of listeners (we refer to this process as ``training the speaker''). We do not have supervision for what to say when playing the game---much as a human player needs to adapt without direct supervision. Instead, we use REINFORCE \citep{reinforce} to train the speaker's adapter parameters, based on the rewards received for differences in the listener accuracies (see Section \ref{sec:multiagentgame}). Our choice to finetune only the adapter builds off prior works in emergent communication that use pretrained-and-frozen  components \citep{dwd, angeliki2020}. By repeated interactions with the same pair of listeners in each experimental condition, the speaker should update its weights to produce specialized language to that listener pair.

\subsection{Listener models}
\label{sec:listeners}
For our listener models (Figure \ref{fig:arch}b,c), we use the vision and language encoders of a pretrained ALIGN NFNet F5 model \citep{align, nfnet}. For a set of input images and caption, we compute image-caption match scores for each image, and the listener selects the image with the highest alignment score.

To investigate the speaker's ability to adapt to distinct listeners, we experimentally manipulate the listeners' knowledge. We use the same pretrained model for each listener, but manipulate each listener's knowledge by applying distinct image transformations to the inputs. For listener 1, we do not transform the input images. For listener 2, we consider the following transformations:\\
\textbf{Crop}: Images are cropped to the top-right quadrant then resized to the original size.\\
\textbf{Blur}: Images are Gaussian blurred (radius 25 px).\\
\textbf{Grayscale}: Images are converted to grayscale.

While listener ``personalities'' could differ along any linguistic, conceptual, and perceptual axes, we focus on these perceptual transformations because they enable quantitative assessments of language specialization. When listener 2 sees cropped images, we expect the speaker to cue objects outside the cropped region. When listener 2 sees blurred images, we expect the speaker to cue objects without the use of color, as distinct objects will not be visible to the second listener but color will still largely be visible. Conversely, when listener 2 sees grayscale, we expect the speaker to only use color words and to specifically stop referring to objects.
We design controlled datasets for each experiment, in which each relevant metric can be measured.

\subsection{Datasets and metrics}
\label{sec:datametrics}

The main datasets we use are transformed versions of Fashion-MNIST \citep{fashionMNIST}. Fashion-MNIST (FMNIST) consists of grayscale images of different types of clothing items. We create a colored version (cFMNIST), where each image is randomly colored with one of 8 colors. To test training with the crop listener, we then create a tiled version of cFMNIST (tcFMNIST) where each image consists of two random cFMNIST images, one in the top-right, and one in the bottom-left. For more realistic images, we use the COCO dataset \citep{coco}, taking center square crops of all images.  Example images for all dataset and transform pairs are shown in Figure \ref{fig:samples}.

For analysis metrics, we first evaluate the proportion of captions generated by the speaker on test images that had color words in them (``color prevalence'') and the proportion of captions that contained a clothing-related word (``FMNIST keyword prevalence''). Each of these metrics identifies language specialization. To measure image-relevance of produced language, we compute metrics measuring how often an object word used corresponded to an actual object in the image, accounting for synonyms.
These metrics are referred to as ``object prevalence'' for cFMNIST, and ``bottom-left object prevalence'' and ``top-right object prevalence'' for tcFMNIST. For colors, it's difficult to enumerate all synonyms as the model often uses pairs of colors, e.g., ``blue and white'' for ``cyan''. Instead, we found that a ``color diversity'' metric---which measures the speaker's variation in color use---offers an effective proxy for determining color relevance.

See Appendices \ref{appx:dataset}, \ref{appx:metrics}, for more details on datasets and metrics, respectively.

\subsection{Training}
\label{sec:training}
A training ``episode'' consists of the speaker receiving an image (observation), generating a caption (action), and receiving the difference in listener accuracies as a reward. As noted above, only the adapter in the speaker has trainable parameters, which are updated via REINFORCE \citep{reinforce}.
We use nucleus sampling \citep{nucleusSampling} to generate captions.
We add a small reward penalty to incentivize short captions (a weight $\lambda$ times the number of words used) as an articulatory effort minimization bias \citep[cf.][]{angelikireview}. We use a batch size of 128 images, with 3 random distractors for each image drawn from the same batch. See Appendix \ref{appx:full-train} for more details.

\section{Results}
\label{sec:results}

\begin{figure}[t]
  \begin{minipage}[t]{\linewidth}
    \includegraphics[width=\linewidth]{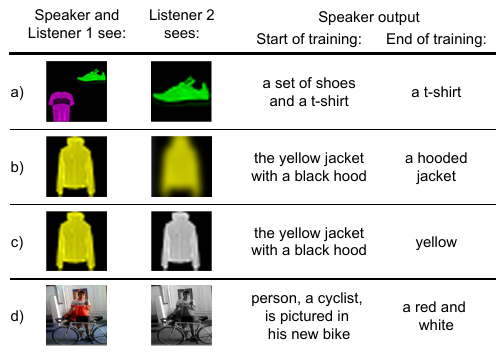}
    \vskip-0.75em
    \caption{Example test images and speaker captions at the start and end of training for various experimental conditions. By row, the speaker trains on a different dataset (a: tcFMNIST, b,c: cFMNIST, d: COCO), and the second listener sees a different transformed image (a: top-right crop, b: blurred, c,d: grayscale).  Diverse language specialization emerges by the end of training. See Appendix \ref{appx:samples} for more samples.}
    \label{fig:samples}
  \end{minipage}
  \vskip-1em
\end{figure}

\subsection{Emergence of specialization} \label{sec:fmnist}

\begin{figure*}[t]
    \centering
    \includegraphics[width=\linewidth]{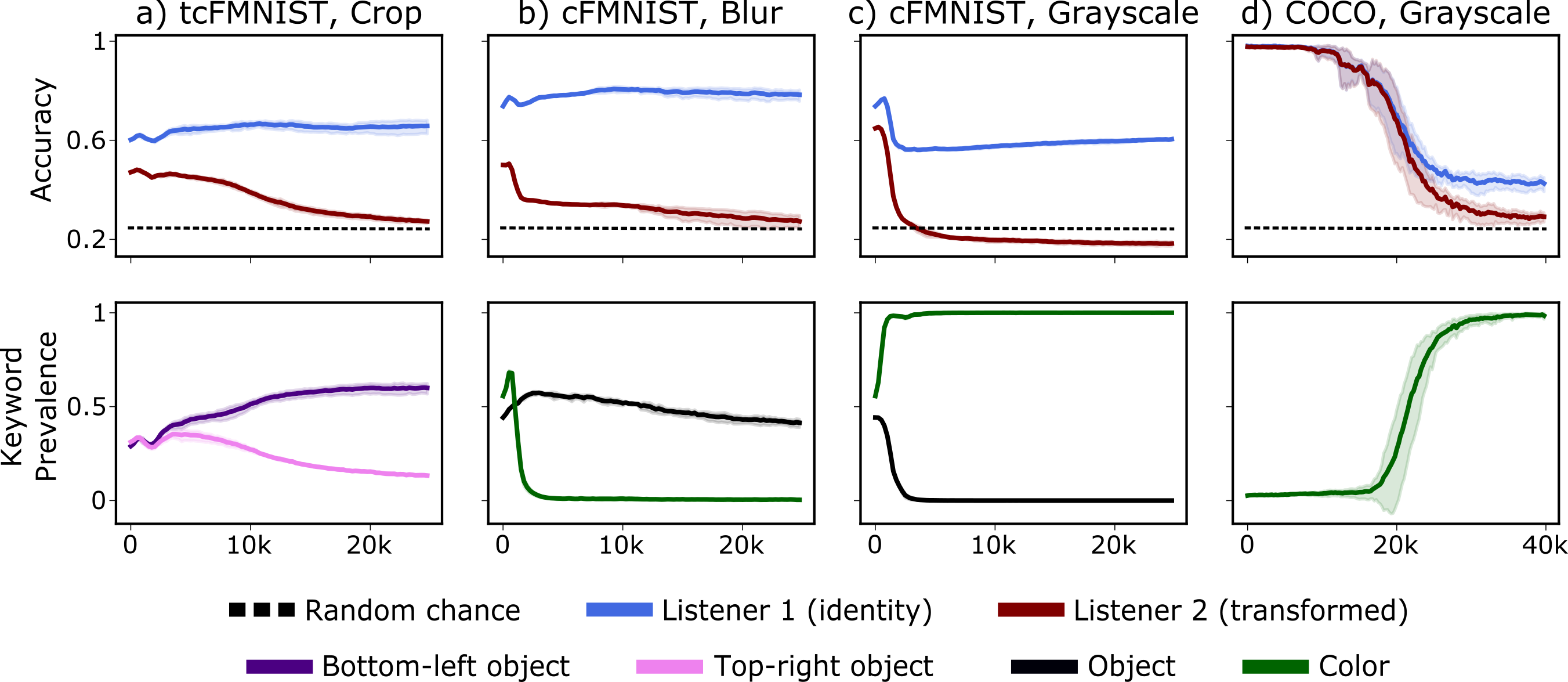}
    \caption{Accuracy and task-relevant metrics for many dataset and listener pairs. By column, the second listener sees a different transformed image (a: top-right crop, b: blurred, c,d: grayscale). The top row shows listener accuracies as the speaker learns to specialize. The bottom row shows prevalance of task-relevant keywords indicated by line color---note that object metrics measure both grounding and language specialization, while color metrics only measure specialization (see Section \ref{sec:datametrics}). Error bars show 95\% confidence intervals (95\% CIs) over 5 seeds.}
    \label{fig:resfig1}
\end{figure*}

By training a speaker to maximize the difference in accuracy between two fixed listeners, we see strong language specialization across many listener pairs (Figure \ref{fig:samples}a-c, \ref{fig:resfig1}a-c). When the second listener sees only the top-right crop of the image, the speaker learns to cue to objects that the second listener cannot see (Figure \ref{fig:resfig1}a, bottom). The first listener's accuracy remains stable through learning, but the second listener's accuracy decreases to chance level. This corresponds to the decrease in top-right object prevalence in the speaker's captions and an \emph{increase} in bottom-left object prevalence. 

When the second listener sees only the blurred version of the image, the speaker learns to reduce its use of color words, while still referring to the target object. The overall learning dynamics are similar to the crop case: the first listener's accuracy remains roughly constant, while the second drops to chance level, again paralleling the decrease in color prevalence in the speaker's captions.

By contrast, when the second listener sees only a grayscale image, we see a rapid decrease in object prevalence and increase in color prevalence. However, this initial learning causes both listener accuracies to drop, as the speaker greedily uses colors that may not be image relevant. Then, from iterations 2.5k to 25k, the speaker slowly learns to use more accurate colors for each image, and the first listener's accuracy increases slowly but steadily (while the second listener stays at chance). At the end of training, color diversity is comparable with that produced by language prompts (see Section \ref{sec:prompting}, Tables \ref{tab:caption_prefix} and \ref{tab:fmnist_color_extended_prefix}), indicating that the speaker is using diverse, image-relevant colors.

These experiments illustrate that the speaker model can optimize its reward in different ways depending on the pair of listeners---decreasing the second listener's accuracy, increasing the first listener's accuracy, or both. Each strategy produces meaningful, measurable changes in the speaker's language, while keeping drift to a minimum. Captions show little structural drift (assessed by an independent language model's likelihoods, Appendix \ref{appx:language_drift}), as well as little semantic and pragmatic drift (as assessed by the above metrics). See Appendix \ref{appx:samples} for more images, distractors, and speaker ouputs. 

\subsection{Extending to real world data}
\label{sec:coco}
We next apply our approach to a setting with real-world data---images from the COCO dataset \citep{coco}. The effect of most transforms is harder to quantify exactly, as the images contain many objects, and enumerating all their synonyms is infeasible. Thus, we use the grayscale transform, which still offers usable metrics (color prevalence and diversity).

Figure \ref{fig:samples}d, \ref{fig:resfig1}d show model samples and the speaker's training curves on COCO. 
\revised{The speaker starts with a high initial accuracy to both listeners. Then, we qualitatively observed (by inspecting model samples) that the speaker starts ``exploring'' different strategies to differentiate the listeners (iterations 0-20k).}
Just before 20k iterations, the speaker starts using colors slightly more often, then exhibits a near-stage-like transition to exclusively using colors.
Qualitatively, the speaker switches standard captions (e.g., ``person in front of a bicycle'') to color-focused captions (e.g., ``red and white'') for each image at some point during training. Once the speaker only uses color for its captions, the individual listener accuracies diverge significantly. Notably both listener accuracies are far worse than at the start of training, as color alone is an imperfect distinguisher on COCO (and overspecifying colors may give away too much, since object category is correlated with color in real images). However, the speaker has successfully learned to specialize its language to exploit the difference between the two listeners. Furthermore, this language remains grounded, as evidenced by end-of-training color diversity being comparable to color diversity produced from language prompting (see Section \ref{sec:prompting}, Tables \ref{tab:caption_prefix} and \ref{tab:coco_extended_prefix}).

\subsection{Zero-shot transfer} \label{sec:zeroshot}
\begin{figure*}[t]
    \centering
    \includegraphics[width=\linewidth]{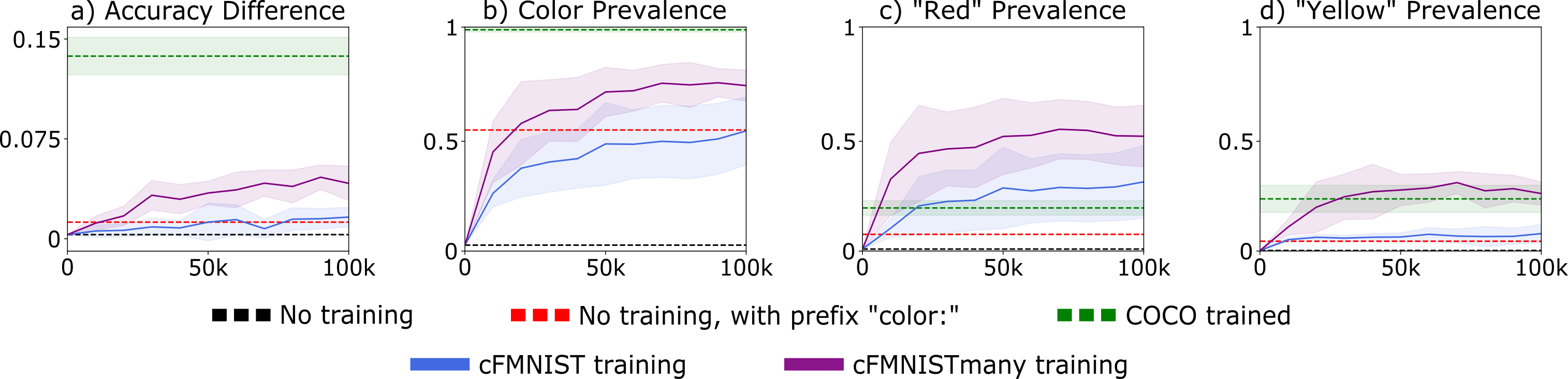}
    \vskip-0.1em
    \caption{Zero-shot transfer results from cFMNIST to COCO and relevant baselines. Each subplot shows a different metric, with different lines corresponding to different conditions or baselines. The two conditions are training on cFMNIST (blue) or cFMNIST with 360 hues (purple). Various baselines are included as dotted lines: base captioning model (black), base captioning model with explicit prompting (red), fully trained on COCO (green).}
    \label{fig:zeroshot}
    \vskip-0.8em
\end{figure*}
We next explore zero-shot transfer of the language specialization from our simpler dataset (cFMNIST) to the more challenging setting of COCO. To do so, we trained a speaker in the grayscale transform setting on cFMNIST (Listener 1 sees the unperturbed image, Listener 2 sees the grayscale image) and tested its zero-shot behavior on COCO. As baselines, we consider the behavior of the pretrained captioning model (that the speaker is initialized to), and the behavior given the best natural-language prompt (see Section \ref{sec:prompting}). We also compare to the behavior of an expert speaker trained to specialize on COCO (the end of training in Figure \ref{fig:resfig1}d).

We find significantly-above-baseline zero-shot transfer, with increasing use of color on COCO as the speaker is trained on FMNIST (Figure \ref{fig:zeroshot}b). Furthermore, this specialization leads to meaningful differences in listener accuracies (Figure \ref{fig:zeroshot}a). This specialization transfer from FMNIST to COCO performs better (across all four metrics in Figure \ref{fig:zeroshot}) than baseline zero-shot approaches (the untuned captioning model), and on par with the best language prompt (see Section \ref{sec:prompting}).

However, zero-shot behavior is far from perfect---the model loses some grounding relative to the experiments above. Zero-shot, the model strongly prefers ``red'' (Figure \ref{fig:zeroshot}c), using it in nearly twice as many captions as the COCO-trained speaker, including even some non-red images. We believe this may be due to the average ``color'' of the 8 colors we use for cFMNIST, in RGB, is 907080, indicating a slight red skew of the data.
To improve generalization, we therefore trained on a richer, balanced colored version of FMNIST, where image hues are sampled randomly from 360 equally-spaced options (as opposed to the 8-hue cFMNIST used above) -- we refer to this dataset as cFMNISTmany. We find that this improves zero-shot transfer across all metrics, pushing it beyond the performance of the best explicit language prompt. Although this model still has slightly increased ``red preference'', its usage of other colors is also significantly better than the 8-hue-trained speaker. For example, in Figure \ref{fig:zeroshot}d, we see that the cFMNISTmany-trained speaker uses yellow in a similar percentage of captions as a COCO-trained speaker. A better captioning model, that was adapted on a more diverse set of images, would likely transfer even better.

\subsection{Ablations}
\label{sec:ablations}
We finally consider ablations to our set-up to identify key aspects. See Appendix \ref{appx:ablations} for full results.

\begin{table*}[t]
    \centering
    \resizebox{\linewidth}{!}{
    \begin{tabular}{@{}c@{}c@{}c@{}c@{}c@{}c@{}c@{}c@{}}
\toprule
\multicolumn{1}{c}{\bfseries Caption Prefix} & \multicolumn{4}{c}{\bfseries FMNIST metrics} & \multicolumn{3}{c}{\bfseries COCO metrics}\\
& \linebreakcell{Accuracy\\Difference} & \linebreakcell{Color\\Prevalence} & \linebreakcell{Color\\Diversity} & \linebreakcell{Object\\ prevalence} & \linebreakcell{Accuracy\\Difference} & \linebreakcell{Color\\Prevalence} & \linebreakcell{Color\\Diversity} \\
\midrule
No prefix & 0.089 & 0.556 & 0.014 & 0.442 & 0.003 & 0.027 & 0.022 \\
the color of this image is & 0.174 & 0.901 & \textbf{0.020} & 0.000 & -0.010 & 0.135 & \textbf{0.036} \\
color: & 0.118 & 0.971 & 0.004 & 0.111 & 0.013 & 0.541 & 0.030\\
colors in image: & 0.048 & 0.999 & 0.001 & 0.000 & 0.019 & 0.335 & 0.018 \\
\midrule
Fully trained & \textbf{0.422} & \textbf{1.000} & 0.015 & \textbf{0.000} & \textbf{0.156} & \textbf{0.988} & 0.033 \\
\bottomrule\\
    \end{tabular}
    }
    \vskip-1.5em
    \caption{Relevant metrics for various caption prefixes.
    The no prefix condition corresponds to the base captioning model. Fully trained results match values at the end of training from Figure \ref{fig:resfig1}---full training via listener subtraction provides superior language specialization compared to explicit prompting.}
    \label{tab:caption_prefix}
    \vskip-1em
\end{table*}

\subsubsection{Single listener}
\label{sec:singlelistener}
In our original experiments, the speaker learns to specialize in the presence of multiple listeners, but are both listeners necessary? Could the speaker also specialize through communicating with only a single listener? To test this possibility, we train a speaker model to optimize the accuracy of just a single listener (so the reward is binary 0/1 listener accuracy). See Appendix \ref{appx:single-train} for details.

In the tcFMNIST case, prevalence of keywords for both locations in the image increases consistently, indicating a lack of specialization to image sub-regions. For cFMNIST, we find that single-listener training leads to increased use of colors and correct keywords; we never see a differential specialization that full multi-listener training induces in our original experiments. For COCO, we see the largest difference between single-listener and multi-listener cases---the single listener case does not learn to use color to distinguish images. 

\subsubsection{Non-contrastive reward}
\label{sec:noncontrastive}
We next consider training with multiple listeners, but a non-contrastive reward. Namely, instead of optimizing for the difference in accuracy between listeners, we optimize directly for the difference in ALIGN match on the cued image. We call this approach ``non-contrastive'' as no distractor images are required. See Appendix \ref{appx:noncontrastive-train} for details.

Training in this setting does lead to some language specialization, but is worse than contrastive training in 3 out of 4 settings we consider. For the crop setting (Table \ref{tab:fmnist_crop_baseline}), we see a reduced use of the top-right keyword, but only a modest increase in the use of the bottom-left keyword compared to contrastive reward training. For the blur setting (Table \ref{tab:fmnist_blur_baseline}), the non-contrastive reward actually outperforms the contrastive reward. The model uses image-relevant keywords 70\% of the time, and stops using varied colors (but continues to mention a single color). For the grayscale setting on both FMNIST and COCO, we observe strong specialization using the non-contrastive reward (as evidenced by low FMNIST keyword prevalence, and high color prevalence in Tables \ref{tab:fmnist_color_baseline}, \ref{tab:coco_baseline}). However, the captions in this case lose image relevance, as evidenced by markedly lower color diversity. Qualitatively, we observe the model using the same color for each image (which achieves more reward than baseline, but is obviously not desirable). Thus, overall, we find that the contrastive reward encourages specialization while preserving image relevance in this multi-listener setting.

\subsubsection{Explicit prompting}
\label{sec:prompting}
Finally, we evaluate a simpler, commonly-used approach to model specialization: explicit prompting. These experiments also provide a baseline for our zero-shot results (in Section \ref{sec:zeroshot}), as no training is involved. We created a variety of possible prefixes to elicit specialization, as explicit prompting can be brittle. Full results are provided in Appendix \ref{appx:prompts}.

Table \ref{tab:caption_prefix} shows the top three prompts and relevant metrics for the (identity, grayscale) transform pair, on both FMNIST and COCO. Explicit prompting does lead to some specialization (when compared to the base captioning model), but fails to provide the robust specialization we observe in Section \ref{sec:results}. Captioning prefixes can be quite brittle, as evidenced by the diverse behavior we observe from superficially similar prompts. For ``colors in image:'' and ``color:'', the model has high color prevalence, but low diversity. Qualitatively, the model tends to list all colors, with only the first color corresponding to the true color of the image. For ``color:'' we also observe that the model still uses nouns. For ``the color of this image is'' we observe slightly reduced color prevalence, but much stronger color diversity and little usage of nouns.

On COCO, similar color prompts yield surprisingly large differences in color prevalence. We also observe ``switching'' behavior: per image, the speaker either ignores the caption prefix (e.g., ``color: person and her son play in the water'') or specializes (e.g., ``color: blue and white''). Qualitatively, this behavior mirrors the stage-like transition in the COCO-trained speaker, in Section \ref{sec:coco}.

\section{Discussion}
\label{sec:discussion}
We have demonstrated a method for specializing a grounded language model without \textit{direct supervision}, by finetuning a small fraction of its parameters in a complex multi-agent setting. We have shown that this approach enables diverse types of specialization with minimal language drift, and have identified the essential aspects of this approach. We believe that our work offers a novel perspective on adapting models to new settings without supervised data.

Our approach to specializing a grounded language model rewards the speaker for using differences in the information it shares with two listener models. This formulation allows for flexible adaptation of models to complex tasks. For example, our approach could potentially be used to diagnose biases in one contrastive listener model by comparing to another (e.g., trained on a different dataset), an area of considerable interest \citep{clipbias}. A speaker optimized for the difference between two contrastive models could identify biases present in one listener, without the need for manually curated word lists.
Additionally, training against listeners of different skill could potentially improve language quality without relying on costly human annotations \cite[e.g.][]{stiennon2020learning}.

Long term, we hope our work contributes to personalized AI. Humans prefer interacting with others who share \textit{rare preferences} \citep{rarepreference}, so an AI that personalizes its language to users would likely be preferred to one that uses standardized language. Personalization could improve other kinds of communication too, like explanations---e.g., adapting feedback to a particular student's idiosyncratic knowledge \citep[cf.][]{studentknowledge}. Our approach enables such adaptation through interaction, without explicit language supervision.

\paragraph{Towards Dixit}
\label{sec:conclusion}
Finally, while our experimental setup is inspired by \textit{Dixit} game, we only make the first steps towards a model that can play the full version of Dixit effectively with humans---something that is an appropriate grand challenge for AI, as \citet{dixitchallenge} suggest. We highlight some of the next steps below.

The present work focuses on perceptual differences, while human play often relies on differences in conceptual knowledge. Finetuning listeners on different datasets (e.g., scenes from Harry Potter movies as in Figure \ref{fig:motivation}a) might allow such conceptual differences to emerge. A challenge would be using datasets where
this conceptual difference is relevant
(Dixit achieves this with abstract images). This could potentially be overcome by generating synthetic data using modern text-to-image models (e.g. \citet{dalle2}). Given the robustness to different listener pairs already exhibited, we believe this is a promising future direction. Moreover, human players adapt their language from a few interactions, while our speaker trains for thousands of iterations. One way to overcome this limitation would be to combine our work with concurrent progress in visual language models. \citet{flamingo} introduce a model, Flamingo, which can rapidly adapt to different visual language tasks in a few-shot setting.
Using such a high-quality base model as the speaker might enable reward-driven adaptation from just a few interactions.

We hope that this work will inspire further research on settings like Dixit, and help to enhance the capabilities of grounded language models and communicative agents more broadly.

\section*{Limitations}
We want to reiterate important limitations with the work in its present form. We only considered settings with two listeners (for ease of quantitative assessment); additional challenges may be present when extending to three or more listeners, which may be necessary to express more complex objectives. In addition, the current work assumes that listeners exist that differ along the axis of desired specialization; finding such listeners might be challenging. Furthermore, despite the overall preservation of language, we qualitatively observed some degradation in some runs.
Introducing a weighted KL-divergence loss term to the pretrained captioning model likelihood \citep[cf.][]{angeliki2020} might further improve specialized language quality. \revised{Alternative sampling methods \citep[cf.][]{decodingComparison}, such as typical decoding \citep{typicalDecoding}, may also improve generated language quality. Finally, our paper does not directly place the agent in interactions with humans, or compare to how humans adapt their language in similar tasks. Further work could include human experiments to get a better sense of language quality and relevance of generated captions and/or identify differences in human and agent language adaptation.}

\section*{Ethical considerations}
\label{sec:ethics}
As discussed above, applications for this work would train speakers using differences between listener models with different weights or even architectures. A potential risk of adapting speakers via this approach is that the speaker might pick up biases that one listener has that are not part of the intended specialization---an undesirable quality. As suggested above, one possible approach to mitigating this would be to explicitly include biased listeners, but always penalize the speaker for the biased listeners' accuracy. Another approach could be debiasing the underlying (frozen) language model \citep{debiasllm, debiasllmreview}, as the speaker's propensity for producing offensive language is inherited from this underlying model. Long-term, we believe that strategies like those detailed by \citet{llmrisk} will be necessary to mitigate risks of large language models (and models such as ours that utilize them).

\section*{Acknowledgements}
The authors would like to thank Frederic Besse, Denis Teplyashin, and Maria Tsimpoukelli for advice on the engineering aspects of the work. We would also like to thank Angeliki Lazaridou, DJ Strouse, Stephanie Chan, and Oriol Vinyals for useful discussions and feedback on the draft.

The authors also thank Flaticon.com for providing free access to various icons used in Figures \ref{fig:motivation} and \ref{fig:arch}. We also thank the creators of Dixit\citep{dixitgame} for creating an inspiring game and the images used in \ref{fig:motivation}.

\bibliographystyle{acl_natbib}
\bibliography{references}

\clearpage

\appendix

\section{Speaker Model Details}
\label{appx:arch-details}
We provide further detail on our speaker model architecture.

The pretrained-and-frozen CLIP visual encoder takes in images of dimension $224 \times 224 \times 3$, which we denote $x$. We take the pre-spatial-pool output from the model, which has dimensions $16 \times 16 \times 768$, and flatten this (across spatial dimensions) to an output $E(x)$, with dimension $256 \times 768$. For the specific encoder model, we experimented with a few (pretrained) options (see Appendix \ref{appx:speaker-pretrain}).

The attention-based (QKV) adapter takes in this visual embedding and transforms it into $n$ token embeddings that can be used as a prefix to prompt the language model. The trainable parameters of this layer are $Q$ (of dimension $n\times d$), $W_K$ (of dimension $768 \times d$), and $W_V$ (of dimension $768 \times 2048$). The weights $W_K$, $W_V$ are used to compute $K = E(x)W_K , V = E(x)W_V$. The fixed queries are then used to attend to these input-dependent keys and values. The output of the adapter layer is thus $A(x) = softmax(QK^\top/\sqrt{d})V$ (of dimension $n \times 2048$).

We feed in $A(x)$ as $n$ prefix embeddings of dimension $2048$ to a pretrained-and-frozen causal Transformer to generate up to 32 tokens (with early termination if the EOS token is produced).
We follow the recommended architecture parameters presented in \citet{chinchilla} for a 1.4B parameter model: our transformer has 24 layers, model dimensionality of $2048$, and $16$ heads. 
We use a SentencePiece tokenizer with a vocabulary size of 32000 \citep{sentencepiece}.

\section{Training details}\label{appx:training}
In this section, we provide training details, hyperparameter search details, and the hyperparameters we used for our final results. All models were implemented in Python using JAX \citep{jax} and Haiku \citep{haiku}. Training was distributed over 16 TPUs (v3), and all experiments used a batch size of 128 (unless specified otherwise). The optimizer for all experiments is Adam \citep{adam}, with $\beta_1 = 0.9, \beta_2 = 0.95$.
We use ZeRO stage-one parameter sharding \citep{zero}.

\subsection{Speaker model pretraining on captioning data}
\label{appx:speaker-pretrain}
We pre-trained our speaker model using supervised cross-entropy loss (with teacher forcing \citep{teacherforcing}) on the Conceptual Captions dataset \citep{conceptualCaptions} which consists of paired image-caption data. 

Dataset images were augmented using random crops. All experiments were run with a batch size of 512 for 500000 steps. Training was distributed over 16 TPUs (v3) in a data parallel fashion.

Hyperparameters we searched over were:
\begin{itemize}
    \item Base CLIP model: We tried the smaller CLIP ViT-B/32 and the larger CLIP ViT-L/14 model.
    \item Positional embeddings: We experimented with adding an absolute positional embedding to each of the $16 \times 16$ unpooled outputs from the CLIP encoder.
    \item Learning rate: [1e-4, 3e-4, 1e-3, 3e-3, 1e-2, 3e-2, 1e-1, 3e-1]
    \item Dimension of QKV adapter ($d$): [32, 64]
    \item Number of input tokens to frozen language model ($n$): [8, 16, 32]
\end{itemize}

All hyperparameter selections were based on validation loss on the Conceptual Captions validation set. We found that the larger CLIP model, no additional positional embeddings, a learning rate of 3e-2, $d=32$, $n=32$, worked best. In terms of the biggest factors, using the larger CLIP model made the biggest change, followed closely by learning rate.

We froze the best model, and that same model served as the starting point for all our experiments.

\revised{
\subsection{Choice to finetune just the adapter}
We experimented with finetuning a visual encoder from scratch (instead of using a pre-trained CLIP) as done by \citet{frozenFSL}. However, we found this approach to perform worse than our adapter-only training (in terms of Conceptual Captions validation performance). The adapter has just under 1.6M parameters, out of a total of 1.7B parameters, for a total of < 0.1\% finetuned parameters. We believe that finetuning such a small number of weights prevents language degeneration, thus enhancing the quality of generated captions. For downstream experiments, we found a similar effect -- finetuning more than just the adapter parameters led to more language drift. For this reason, we only finetune the adapter in all experiments.
}

\subsection{Multi-listener, contrastive-reward training}
\label{appx:full-train}

\begin{table*}[ht]
    \centering
\begin{tabular}{c c c c c}
\toprule
\multicolumn{1}{c}{} & \multicolumn{3}{c}{\bfseries FMNIST} & \multicolumn{1}{c}{\bfseries COCO}\\
    Parameter & Crop & Blur & Grayscale & COCO Grayscale  \\
    \hline
    Learning rate & 3e-4 & 3e-4 & 3e-4 & 3e-4 \\
    Sampling temperature & 1 & 1 & 1 & 1 \\
    Nucleus Size & 0.8 & 0.8 & 1 & 1 \\
    Caption length penalty $(\lambda)$ & 3e-3 & 3e-3 & 3e-3 & 3e-4 \\
    \bottomrule
\end{tabular}
    \caption{Hyperparameters for the main setting (multi-listener, contrastive-reward).}
    \label{tab:main-params}
\end{table*}

We used the hyperparameters detailed in Table \ref{tab:main-params} for training our main models. These hyperparameters were found by grid searches over learning rate ([3e-5, 3e-4]), sampling temperature ([1,2]), nucleus size ([0.8, 1]), and caption length penalty ([1e-5, 3e-5, 1e-4, 3e-4, 1e-3, 3e-3, 1e-2, 3e-2]). We found that caption length penalty was crucial to tune to make sure it didn't dominate the reward at the start of training. 

All of our final training results are run on 5 different random seeds, from which 95\% confidence intervals are calculated and shown (in figures and tables).

\subsection{Single listener, contrastive-reward training}
\label{appx:single-train}

\begin{table*}[ht]
    \centering
\begin{tabular}{c c c c}
    \toprule
    Parameter & cFMNIST & tcFMNIST & COCO  \\
    \hline
    Learning rate & 3e-4 & 3e-4 & 3e-4 \\
    Sampling temperature & 1 & 1 & 1 \\
    Nucleus Size & 0.8 & 0.8 & 1 \\
    Caption length penalty $(\lambda)$ & 1e-5 & 1e-5 & 1e-5 \\
    \bottomrule
\end{tabular}

    \caption{Hyperparameters for the single listener, contrastive-reward baseline.}
     \label{tab:single-params}
\end{table*}
For this baseline, we train the speaker to just maximize the binary accuracy of the single unperturbed listener. We train in 3 settings: on cFMNIST, on tcFMNIST, and on COCO. To make a fair comparison, we performed a hyperparameter search for each case (final hyperparameters shown in Table \ref{tab:single-params}). Specifically, we searched over learning rate ([3e-5, 3e-4]), nucleus size ([0.8, 1]), and caption length penalty ([1e-5, 1e-4, 1e-3, 1e-2]). We also found that early stopping and batch-based baseline subtraction \citep{reinforce} improved performance (in terms of reward on unseen images), so we used both of these techniques for all results reported. On each setting, we trained 5 random seeds for up to 25000 iterations on FMNIST and up to 40000 iterations on COCO, early stopped each one, and computed evaluation metrics on those checkpoints.

\subsection{Multi listener, non-contrastive-reward training}
\label{appx:noncontrastive-train}
\begin{table*}[ht]
    \centering
\begin{tabular}{c c c c c}
\toprule
\multicolumn{1}{c}{} & \multicolumn{3}{c}{\bfseries FMNIST} & \multicolumn{1}{c}{\bfseries COCO}\\
    Parameter & Crop & Blur & Grayscale & COCO Grayscale  \\
    \hline
    Learning rate & 3e-4 & 3e-4 & 3e-4 & 3e-4 \\
    Sampling temperature & 1 & 1 & 1 & 1 \\
    Nucleus Size & 0.8 & 0.8 & 0.8 & 0.8 \\
    Caption length penalty $(\lambda)$ & 3e-6 & 3e-6 & 3e-6 & 3e-4 \\
    \bottomrule
\end{tabular}
        \caption{Hyperparameters for the multi-listener, non-contrastive-reward baseline.}
         \label{tab:noncontrastive-params}
\end{table*}
For this baseline, we train the speaker to just maximize the difference in the ALIGN match scores that listeners assign to the cued image. In this setup, the listeners only need the caption and cued image, which is why we call it non-contrastive. We train this baseline in all four settings that we train our main (multi-listener, contrastive-reward) model in. We perform hyperparameter over nucleus size ([0.8, 1]) and caption length penalty ([1e-7, 1e-6, 3e-6, 1e-5, 3e-5, 1e-4, 3e-4, 1e-3]), with optimal hyperparameters reported in Table \ref{tab:noncontrastive-params}.

\section{Dataset details}
\label{appx:dataset}

\begin{table*}[ht]
    \centering
    \begin{tabular}{c c c c c c c c c}
    \toprule
        {\bfseries Color:} &  Red & Orange & Yellow & Green & Cyan & Blue & Purple & Pink\\
        {\bfseries Hue:} & $\frac{0}{360}$ & $\frac{30}{360}$ & $\frac{60}{360}$ & $\frac{120}{360}$ & $\frac{180}{360}$ & $\frac{240}{360}$ & $\frac{270}{360}$ & $\frac{300}{360}$ \\
        \bottomrule
    \end{tabular}
    \caption{Color and Hues for cFMNIST.}
    \label{tab:dataset-colors}
\end{table*}
To construct the cFMNIST dataset, we convert FMNIST images to HSV. Since the images are originally grayscale, the saturation is always 0. We set the saturation to 1, and set the hue to one of the 8 colors shown in Table \ref{tab:dataset-colors}. Then, we convert back to RGB images and resize to spatial dimensions $224 \times 224$  using cubic upsampling \citep{bicubic}, then clip all values to the range [0,1]. 

For tcFMNIST, we randomly place a cFMNIST image in the bottom left and in the top-right of the image. This choice of tiling (as opposed to top-left, bottom-right, or some other combination) was chosen to avoid an inherent bias towards cuing for the top-left object that we observed in our base captioning model.

For the test set, we construct a fixed test set to use for equal comparison across all conditions. For contrastive-reward experiments, we also fix the distractors during test time for consistency.

We normalize all images according to the normalization that was originally used for the off-the-shelf image encoders we use (for consistency). Specifically, for inputs to the speaker, which pass through CLIP's vision encoder, the normalization mean and standard deviation values are (0.481, 0.458, 0.408) and (0.269, 0.261, 0.278), respectively. For inputs to the listeners, which pass through ALIGN's vision encoder, the normalization mean and standard deviation values are (0.485, 0.456, 0.406) and (0.229, 0.224, 0.225), respectively.

\subsection{Licenses}
\label{sec:license}
We use FMNIST under the MIT license, COCO under the creative commons 4.0 attribution license, and Conceptual Captions under its ad-hoc license (for which we thank Google LLC). We also acknowledge that images in Figure \ref{fig:motivation} are taken from the Dixit game \citep{dixitgame} and icons are used under the Flaticon license (for which we thank Flaticon.com). We make use of the GPT3 \citep{gpt3} beta (for quantification of structural drift, see Appendix \ref{appx:language_drift}) under the Apache License.

To our knowledge, we use these resources in agreement with their licenses.

\subsection{Dataset Splits}
\label{sec:splits}
We use standard splits for all datasets from tensorflow datasets \cite{tensorflow2015-whitepaper}. For FMNIST-based datasets, this means 60000 train images that are used to procedurally generate cFMNIST and tcFMNIST train sets, and 10000 test images that are used to generate the test sets. 

\section{Evaluation metric details}
\label{appx:metrics}

We have two types of metrics: color metrics and keyword metrics. Color metrics apply on both FMNIST-based datasets and COCO, while keyword metrics are only used on FMNIST-based dataset (where we know exactly where and what the objects are by construction). We want our metrics to be able to diagnose language specialization, and also check that the captions are still grounded (we'll refer to this as ``image relevance'' of captions).

\subsection{Color metrics}
All color metrics utilize the following set of 16 colors: 
\begin{verbatim}
    red, orange, yellow, green, 
    blue, indigo, violet, purple, 
    cyan, magenta, pink, brown, 
    black, white, gray, grey
\end{verbatim}
For measuring language specialization to colors, we define \textit{color prevalence}: the fraction of captions at test time containing at least one word from the above list.

For measuring image relevance, we use a proxy metric as its often hard to define what it means for a color to be correct (e.g., if an image is ``cyan'' and the model says ``blue and white''). The proxy metric we use is \textit{color diversity}. To compute color diversity we calculated TF-IDF vectors for each caption using only the color terms above. Then, we calculated the trace of the covariance matrix (which has dimension $16\times 16$) of these vectors. This metric has the desirable property where if a color appears in nearly all captions, it will be scaled down by the IDF term and so will its contribution to the trace of the covariance matrix. 

We found this to be a good proxy metric as loss of image relevance in color specialization cases corresponding to the model choosing just a few colors (often just one) and using them to describe all images. To not confound our main results, we qualitatively looked at how well this metric corresponded to loss of image relevance on the many language prompts detailed in Appendix \ref{appx:prompts}. We found that color diversity was a noisy, but useful, metric for determining image relevance of these prompts. For example, a simpler metric like checking the number of unique captions the model uses fails since, given some language prompts, the speaker would produce long strings of colors, often in slightly different orders (e.g. ``color: green, black, white, blue, red ...'' vs. ``color: red, black, white, blue, green ...''). Furthermore, we note that color diversity is not a perfect metric, so subtle differences (on the order of $\pm$ 0.005) should not be considered significant. We mainly use color diversity to classify runs where image relevance was completely lost (color diversity going to 0), or image relevance was retained (color diversity staying near that of the best language prompts and above that at the start of training).

\subsection{Keyword metrics}
\begin{table*}[ht]
    \centering
    \begin{tabular}{c c c}
    \toprule
        Label & Category name & Added synonyms \\
        \hline
        0 & t-shirt & top, t-shirts, shirt, shirts  \\
        1 & trouser & trousers, pants \\
        2 & pullover & sweater, hoodie, sweaters, hoodies \\
        3 & dress & dresses \\
        4 & coat & coats, jacket, jackets \\
        5 & sandal & high heels, heels, shoe, shoes \\
        6 & shirt & shirts \\
        7 & sneaker & sneakers, shoe, shoes, running shoe \\
        8 & bag & purse, backpack, bags, purses \\
        9 & ankle boot & boot, shoe, shoes, boots \\
        \hline
    \end{tabular}
    \caption{FMNIST keyword sets. We refer to the set of all words in this table as FMNIST related keywords.}
    \label{tab:keywords}
\end{table*}
For measuring language specialization to objects, we define \textit{FMNIST keyword prevalence}: the fraction of captions at test time containing at least one FMNIST related keyword.

To measure image relevance, we utilize the ground-truth labels from FMNIST, supplemented with synonyms. Specifically, we measure \textit{object prevalence}: the fraction of captions at test time that contain at least one keyword corresponding to the ground truth label of the object in, the image according to Table \ref{tab:keywords}. We found that allowing synonyms was essential, as the captioning model heavily prefers some clothing words over others (e.g., sandals are almost always just called shoes by the model). For tcFMNIST, we similarly define \textit{bottom-left object prevalence} and \textit{top-right object prevalence} to measure what fraction of captions refer to each region of the image.

\section{Language drift quantification} \label{appx:language_drift}
Prior work in emergent communication \citep{angeliki2020} establishes three types of language drift that may occur when adapting language from rewards, without direct supervision: structural drift (how ``language-like'' are captions), semantic drift (how ``image-relevant'' are captions), and pragmatic drift (how ``human-interpretable'' are captions). While the focus of our work is on language specialization, it is important to investigate to what extent our approach is resulting in language drift. 

For semantic and pragmatic drift, we note that our main metrics (see Section \ref{sec:datametrics} and Appendix \ref{appx:metrics}) measure human interpretability as well as image relevance. For example, our object metrics measure whether the model is referring to objects using the category name or valid synonyms, and our color diversity metric is able to differentiate settings with notable semantic drift (e.g., the non-contrastive baseline, see Section \ref{sec:noncontrastive}) from settings without notable semantic drift (e.g., our main results, see Figures \ref{fig:samples} and \ref{fig:resfig1}). Our method preserves these metrics, presumably since the grounded task with frozen listeners, as well as the frozen components of the speaker, provide strong constraints on what language the speaker can use.

However, our main metrics do not adequately address the issue of structural drift. To measure how ``language-like'' our captions are, we therefore follow the approach of \citet{angeliki2020} and evaluate the log-likelihood assigned to generated captions by an independent, pretrained language model (LM). Specifically, we use OpenAI's Ada model, made available online through the GPT3 beta \citep{gpt3}. For comparison, we also show LM log-likelihoods for the ground truth human captions (for cFMNIST, we procedurally generate these---e.g., ``red pullover''), the LM likelihoods for the ``best prompt'' (see Appendix \ref{appx:prompts}), and the LM likelihoods for captions from the speaker before specialization training (the base speaker pretrained on captioning only). For reference, the ``best'' language prompts were chosen to maximize task-relevant metrics (see third column in Tables \ref{tab:fmnist_crop_extended_prefix}-\ref{tab:coco_extended_prefix}) except for the (cFMNIST, Grayscale) case where we use the third best (as the top two prefixes have very low color diversity---see discussion in Section \ref{sec:prompting}).
Results are shown in Table \ref{tab:lm_full}.

\begin{table*}[h]
    \centering
    \resizebox{\textwidth}{!}{
    \begin{tabular}{c c c c c}
    \toprule
         & tcFMNIST, Crop & cFMNIST, Blur & cFMNIST, Grayscale & COCO, Grayscale \\
         \hline
        Ground truth & -32.77 & \textbf{-15.20} & -15.20 & -46.26 \\
        ``Best'' language prompt & -36.61 & -44.61 & -24.59 & -29.49 \\
        Start of training & -33.33 & -28.64 & -28.64 & \textbf{-28.44} \\
        End of training & \textbf{-17.41} & -19.01 & \textbf{-9.26} & -29.49 \\
        \bottomrule
    \end{tabular}
    }
    \caption{Average language model \emph{full-caption} log-likelihoods for ground truth, start-of-training, and end-of-training captions on test images.}
    \label{tab:lm_full}
\end{table*}

\begin{table*}[h]
    \centering
    \resizebox{\textwidth}{!}{
    \begin{tabular}{c c c c c}
    \toprule
         & tcFMNIST, Crop & cFMNIST, Blur & cFMNIST, Grayscale & COCO, Grayscale \\
         \hline
        Ground truth & -5.823 & -7.666 & -7.666 & -4.772 \\
        ``Best'' language prompt & -4.680 & -4.371 & \textbf{-4.076} & -4.051 \\
        Start of training & \textbf{-4.160} & \textbf{-4.127} & -4.127 & -4.463 \\
        End of training & -7.143 & -8.383 & -4.234 & \textbf{-2.509} \\
        \bottomrule
    \end{tabular}
    }
    \caption{Average language model \emph{per-token} log-likelihoods for ground truth, start-of-training, and end-of-training captions on test images.}
    \label{tab:lm_per_token}
\end{table*}

Surprisingly, we see that in most of our runs, average caption likelihoods seem to increase as the speaker specializes. It appears that this effect may be driven by the length penalty---over training, the speaker produces shorter captions, which have higher likelihoods since they have fewer tokens. Qualitatively, we find that language drift can be fairly variable across different random seeds. For example, the COCO average is worse after training largely due to a single run, in which long repeated captions emerged (which have lower likelihoods than their un-repeated counterparts).

To evaluate this further, we computed the per-token likelihoods (Table \ref{tab:lm_per_token}). While these do show some decrease in likelihood in some cases, the captions at the end of training are generally of comparable likelihood to the ground truth. In this instance, we notice that the one COCO seed in which the captions repeated actually exhibits \emph{greater} per-token likelihood---after a few repeats, the LM starts to estimate further repeats to be very likely. This could potentially be a concern for using LM likelihoods as a metric for structural drift more broadly \citep[cf.][]{counteringdrift}. However, in most runs our length penalty prevents repetitions or long captions, as noted above.

In summary, these results indicate that language has not drifted far in most conditions and for most random seeds. We attribute this to some of the same factors that help prevent other types of language drift: finetuning a small part of our speaker (just the adapter), keeping the listener models frozen (thus avoiding co-adaptation), using a length penalty (see above discussion), and using contrastive reward (crucial for combating semantic drift, see Section \ref{sec:noncontrastive}). If necessary, language drift could potentially be reduced further  by using a KL-divergence loss to the distribution of outputs from the base captioning model (as done by \citet{angeliki2020}).

\section{Full ablation results} \label{appx:ablations}

In this section we show the full quantitative ablation results, in Tables \ref{tab:fmnist_crop_baseline}, \ref{tab:fmnist_blur_baseline}, \ref{tab:fmnist_color_baseline}, and \ref{tab:coco_baseline}.

\begin{table*}[ht]
    \centering
    \begin{tabular}{@{}c@{}c@{}c@{}}
\toprule
 Condition & \linebreakcell{Bottom-left object\\prevalence} & \linebreakcell{Top-right object\\prevalence} \\ \midrule
No training & 0.290 & 0.312 \\
Single listener & $0.475 \pm 0.024$ & $0.473 \pm 0.020$ \\
Non-contrastive & $0.360 \pm 0.005$ & $\mathbf{0.120 \pm 0.011}$ \\
Full training & $\mathbf{0.602 \pm 0.066}$ & $0.135 \pm 0.017$ \\
\bottomrule\\
    \end{tabular}
    \caption{Metrics (with 95\% CI) for ablations on the tcFMNIST dataset and the (identity, crop) pair of listener transformations. Both non-contrastive and full training lead to decreased use of top-right keyword, but only full training accurately specializes to using the bottom-left keyword.}
    \label{tab:fmnist_crop_baseline}
\end{table*}

\begin{table*}[ht]
    \centering
    \begin{tabular}{ccccc}
 \toprule
 Condition &  \linebreakcell{FMNIST\\keyword\\prevalence} & \linebreakcell{Object\\prevalence} & \linebreakcell{Color\\prevalence} & \linebreakcell{Color\\diversity} \\
 \midrule
No training & 0.644 & 0.442 & 0.556 & 0.014 \\
Single listener & $0.754 \pm 0.020$ & $0.564 \pm 0.013$ & $0.953 \pm 0.022$ & $0.016 \pm 0.000$ \\
Non-contrastive & $\mathbf{0.997 \pm 0.004}$ & $\mathbf{0.710 \pm 0.086}$ & $0.606 \pm 0.833$ & $\mathbf{0.000 \pm 0.000}$ \\
Full training & $0.600 \pm 0.044$ & $0.418 \pm 0.038$ & $\mathbf{0.004 \pm 0.008}$ & $0.001 \pm 0.002$ \\
    \bottomrule\\
    \end{tabular}
    \caption{Relevant metrics (with 95\% CI) for various ablations on the cFMNIST dataset and the (identity, blur) pair of listener transformations. Non-contrastive rewards perform better than contrastive rewards in this setting, as evidenced by increased use of correct keywords, and decreased use of meaningful colors (diversity of colors goes to 0).}
    \label{tab:fmnist_blur_baseline}
\end{table*}

\begin{table*}[ht]
    \centering
    \begin{tabular}{ccccc}
    \toprule
 Condition & \linebreakcell{Color\\prevalence}& \linebreakcell{Color\\diversity} & \linebreakcell{FMNIST\\keyword\\prevalence} & \linebreakcell{Object\\prevalence} \\
 \midrule
No training & 0.556 & 0.014 & 0.644 & 0.442 \\
Single listener & $0.953 \pm 0.022$ & $\mathbf{0.016 \pm 0.000}$ & $0.754 \pm 0.020$ & $0.564 \pm 0.013$ \\
Non-contrastive & $1.000 \pm 0.000$ & $0.005 \pm 0.003$ & $0.000 \pm 0.000$ & $0.000 \pm 0.000$ \\
Full training & $\mathbf{1.000 \pm 0.000}$ & $0.015 \pm 0.002$ & $\mathbf{0.000 \pm 0.000}$ & $\mathbf{0.000 \pm 0.000}$ \\
\bottomrule\\
    \end{tabular}
    \caption{Relevant metrics (with 95\% CI) for various ablations on the cFMNIST dataset and the (identity, grayscale) pair of listener transformations. Full training performs best as color prevalence and diversity increase, while keyword prevalence decreases to 0.}
    \label{tab:fmnist_color_baseline}
\end{table*}

\begin{table*}[ht]
    \centering
    \begin{tabular}{@{}ccc@{}}
    \toprule
 & Color prevalence & Color diversity \\
 \midrule
No training & 0.027 & 0.022 \\
Single listener & $0.102 \pm 0.080$ & $0.030 \pm 0.007$ \\
Non-contrastive & $0.932 \pm 0.256$ & $0.018 \pm 0.005$ \\
Full training & $\mathbf{0.988 \pm 0.023}$ & $\mathbf{0.033 \pm 0.027}$ \\
\bottomrule \\
    \end{tabular}
    \caption{Relevant metrics (with 95\% CI) for various ablations on the COCO dataset and the (identity, grayscale) pair of listener transformations. Full training performs best as it has highest color prevalence and diversity.}
    \label{tab:coco_baseline}
\end{table*}
\clearpage
\section{Extended results on caption prefixes}
\label{appx:prompts}

In Tables \ref{tab:fmnist_crop_extended_prefix}-\ref{tab:coco_extended_prefix}, we show all caption prefixes we experimented with, as well as the relevant metrics for each. For each table, we sort prefixes from worst (top of table) to best (bottom of table) based on the most task-relevant keyword metric (third column in each table). We added a column for ``Score difference'' which is the average difference in ALIGN match scores from each listener for the cued image (it corresponds to the reward that is seen in the non-contrastive reward baseline). Of course, no training occurs, as the caption prefixes just explore how well the speaker can do by just explicit prompting. We also use \textvisiblespace to indicate a space at the end of the caption. For most prefixes, we see a large difference between adding this space and not adding it, which is just another testament to how brittle prompt engineering can be.

\begin{table*}[h]
    \centering
    \resizebox{\textwidth}{!}{
    \begin{tabular}{c p{1.5cm} p{1.4cm} p{1.4cm} p{1.4cm} p{1.4cm}}
\toprule
Caption prefix & Accuracy difference & Score difference & Bottom-left object prevalence & Top-right object prevalence & FMNIST keyword prevalence \\
\hline
the bottom left of this picture is\textvisiblespace & 0.0121 & 0.0061 & 0.0181 & 0.0193 & 0.0502 \\
the bottom left of this image is\textvisiblespace & 0.0052 & 0.0065 & 0.0241 & 0.0277 & 0.0794 \\
in the bottom left of this image, there is\textvisiblespace & 0.0512 & 0.0112 & 0.0663 & 0.0719 & 0.1688 \\
bottom left of this image:\textvisiblespace & 0.0155 & 0.0096 & 0.0711 & 0.0798 & 0.2604 \\
the bottom left of this image shows\textvisiblespace & 0.0338 & 0.0137 & 0.0993 & 0.1212 & 0.3179 \\
bottom left:\textvisiblespace & 0.0418 & 0.0439 & 0.1154 & 0.1274 & 0.4156 \\
in the bottom left of this image, there is & 0.0681 & 0.0262 & 0.2711 & 0.2954 & 0.6176 \\
the bottom left of this image is & 0.1085 & 0.0293 & 0.3027 & 0.3193 & 0.8302 \\
the bottom left of this picture is & 0.1067 & 0.0365 & 0.3189 & 0.3420 & 0.8320 \\
the bottom left of this image shows & 0.1256 & 0.0515 & 0.3428 & 0.3342 & 0.6091 \\
bottom left of this image: & 0.1113 & 0.0340 & 0.3561 & 0.3772 & 0.7671 \\
bottom left: & 0.1093 & 0.0702 & 0.3597 & 0.3770 & 0.7886 \\
\bottomrule
    \end{tabular}}
    \caption{Various metrics for prompted generation of base captioning model on tcFMNIST. Accuracy and score difference are calculated on the (unperturbed, crop) pair of listeners.}
    \label{tab:fmnist_crop_extended_prefix}
\end{table*}

\begin{table*}[h]
    \centering
    \resizebox{\textwidth}{!}{
    \begin{tabular}{c p{1.5cm} p{1.4cm} p{1.4cm} p{1.4cm} p{1.4cm} p{1.4cm}}
\toprule
Caption prefix & Accuracy difference & Score difference & Object prevalence & FMNIST keyword prevalence & Color prevalence & Color diversity \\
\hline
the clothing item in this image is\textvisiblespace & 0.0174 & 0.0491 & 0.0685 & 0.1390 & 0.0801 & 0.0083 \\
an image of an object:\textvisiblespace & 0.0265 & 0.0060 & 0.0743 & 0.0981 & 0.0755 & 0.0103 \\
a black and white image of\textvisiblespace & 0.0493 & -0.0071 & 0.1065 & 0.1439 & 1.0000 & 0.0044 \\
the item in this image is\textvisiblespace & 0.0335 & 0.0254 & 0.1147 & 0.1493 & 0.0532 & 0.0083 \\
the object in this image is\textvisiblespace & 0.0361 & -0.0008 & 0.1321 & 0.1664 & 0.0791 & 0.0104 \\
item:\textvisiblespace & 0.0280 & 0.0433 & 0.1627 & 0.2210 & 0.3257 & 0.0131 \\
object:\textvisiblespace & 0.0558 & -0.0017 & 0.2274 & 0.3169 & 0.3764 & 0.0141 \\
a picture of\textvisiblespace & 0.0664 & 0.0044 & 0.2350 & 0.3213 & 0.4122 & 0.0155 \\
an image of\textvisiblespace & 0.0824 & 0.0103 & 0.2646 & 0.3684 & 0.4778 & 0.0157 \\
a black and white image of & 0.1170 & 0.0362 & 0.3895 & 0.5977 & 1.0000 & 0.0050 \\
the item in this image is & 0.0988 & 0.0449 & 0.4014 & 0.5656 & 0.3498 & 0.0136 \\
a picture of & 0.0997 & 0.0518 & 0.4363 & 0.6556 & 0.4765 & 0.0165 \\
an image of & 0.1059 & 0.0466 & 0.4428 & 0.6638 & 0.5181 & 0.0163 \\
the object in this image is & 0.0885 & 0.0285 & 0.4438 & 0.6146 & 0.4448 & 0.0155 \\
an image of an object: & 0.0967 & 0.0363 & 0.4523 & 0.6343 & 0.4434 & 0.0156 \\
object: & 0.1183 & 0.0018 & 0.4545 & 0.6533 & 0.5643 & 0.0143 \\
item: & 0.1015 & 0.0528 & 0.4620 & 0.6731 & 0.5747 & 0.0134 \\
the clothing item in this image is & 0.1086 & 0.0546 & 0.4692 & 0.8643 & 0.5157 & 0.0142 \\
\bottomrule
    \end{tabular}}
\caption{Various metrics for prompted generation of base captioning model on cFMNIST. Accuracy and score difference are calculated on the (unperturbed, blur) pair of listeners.}
\label{tab:fmnist_blur_extended_prefix}
\end{table*}

\begin{table*}[h]
    \centering
    \resizebox{\textwidth}{!}{
    \begin{tabular}{c p{1.5cm} p{1.4cm} p{1.4cm} p{1.4cm} p{1.4cm} p{1.4cm}}
\toprule
Caption prefix & Accuracy difference & Score difference & Color prevalence & Color diversity & Object prevalence & FMNIST keyword prevalence \\
\hline
color color color:\textvisiblespace & 0.0180 & 0.0264 & 0.1421 & 0.0055 & 0.0330 & 0.0433 \\
the colors in this image are\textvisiblespace & 0.0139 & 0.0341 & 0.1697 & 0.0082 & 0.0040 & 0.0045 \\
a picture with the color\textvisiblespace & 0.0584 & 0.0410 & 0.2206 & 0.0139 & 0.2622 & 0.3536 \\
a picture with the color & 0.0851 & 0.0507 & 0.2745 & 0.0144 & 0.3423 & 0.4613 \\
a picture with color\textvisiblespace & 0.0347 & 0.0395 & 0.2891 & 0.0141 & 0.1583 & 0.2319 \\
color:\textvisiblespace & 0.0507 & 0.0320 & 0.3000 & 0.0060 & 0.0387 & 0.0501 \\
the color of this image is\textvisiblespace & 0.0318 & 0.0341 & 0.3648 & 0.0145 & 0.0058 & 0.0064 \\
an image with the color\textvisiblespace & 0.0782 & 0.0302 & 0.3862 & 0.0153 & 0.2468 & 0.3197 \\
an image with the color & 0.0952 & 0.0553 & 0.3905 & 0.0153 & 0.2743 & 0.3579 \\
an image with color\textvisiblespace & 0.0268 & 0.0308 & 0.4225 & 0.0136 & 0.1661 & 0.2413 \\
the colors in this image are & 0.0587 & 0.0511 & 0.4336 & 0.0063 & 0.0490 & 0.0547 \\
a picture with color & 0.0452 & 0.0398 & 0.4396 & 0.0111 & 0.1848 & 0.2455 \\
colors in image:\textvisiblespace & 0.0097 & 0.0397 & 0.5018 & 0.0020 & 0.0002 & 0.0004 \\
color color color: & 0.0764 & 0.0572 & 0.5512 & 0.0033 & 0.1996 & 0.2535 \\
an image colored\textvisiblespace & 0.0919 & 0.0426 & 0.5901 & 0.0123 & 0.1259 & 0.1593 \\
an image with color & 0.0504 & 0.0410 & 0.6026 & 0.0112 & 0.1863 & 0.2555 \\
an image colored & 0.1303 & 0.0631 & 0.8600 & 0.0101 & 0.0200 & 0.0271 \\
the color of this image is & 0.1735 & 0.0667 & 0.9010 & 0.0198 & 0.0002 & 0.0002 \\
color: & 0.1180 & 0.0583 & 0.9705 & 0.0040 & 0.1110 & 0.1354 \\
colors in image: & 0.0477 & 0.0428 & 0.9994 & 0.0012 & 0.0000 & 0.0000 \\
\bottomrule
    \end{tabular}}
 \caption{Various metrics for prompted generation of base captioning model on cFMNIST. Accuracy and score difference are calculated on the (identity, grayscale) pair of listeners. For reference, the color diversity of the fully specialized speaker in this case is 0.015, which is on par with the best color diversity values across prompts.}
 \label{tab:fmnist_color_extended_prefix}
\end{table*}

\begin{table*}[h]
    \centering
    \resizebox{\textwidth}{!}{
    \begin{tabular}{c c c c c}
\toprule
Caption prefix & Accuracy difference & Score difference & Color prevalence & Color diversity \\
\hline
an image colored & 0.0016 & 0.0253 & 0.0109 & 0.0110 \\
a picture with the color & 0.0055 & 0.0136 & 0.0117 & 0.0144 \\
a picture with color\textvisiblespace & -0.0039 & 0.0023 & 0.0172 & 0.0155 \\
the color of this image is\textvisiblespace & -0.0094 & -0.0071 & 0.0195 & 0.0180 \\
color color color:\textvisiblespace & -0.0086 & 0.0045 & 0.0219 & 0.0172 \\
an image colored\textvisiblespace & -0.0195 & 0.0166 & 0.0242 & 0.0190 \\
an image with color\textvisiblespace & -0.0141 & -0.0010 & 0.0242 & 0.0182 \\
the colors in this image are & -0.0008 & 0.0131 & 0.0312 & 0.0171 \\
colors in image:\textvisiblespace & -0.0055 & -0.0023 & 0.0344 & 0.0128 \\
an image with the color & 0.0000 & 0.0115 & 0.0375 & 0.0221 \\
a picture with the color\textvisiblespace & 0.0031 & 0.0040 & 0.0383 & 0.0223 \\
color:\textvisiblespace & 0.0078 & 0.0079 & 0.0508 & 0.0224 \\
an image with the color\textvisiblespace & 0.0055 & -0.0021 & 0.0586 & 0.0261 \\
color color color: & -0.0016 & 0.0071 & 0.0906 & 0.0278 \\
a picture with color & -0.0016 & 0.0053 & 0.0914 & 0.0165 \\
the colors in this image are\textvisiblespace & -0.0156 & 0.0091 & 0.1172 & 0.0317 \\
the color of this image is & -0.0102 & 0.0012 & 0.1352 & 0.0362 \\
an image with color & 0.0008 & 0.0029 & 0.1477 & 0.0167 \\
colors in image: & 0.0188 & 0.0052 & 0.3352 & 0.0179 \\
color: & 0.0125 & 0.0047 & 0.5406 & 0.0304 \\
\bottomrule
    \end{tabular}}
\caption{Various metrics for prompted generation of base captioning model on COCO. Accuracy and score difference are calculated on the (identity, grayscale) pair of listeners. For reference, the color diversity of the fully specialized speaker in this case is 0.033, which is on par with the best color diversity values across prompts.}
\label{tab:coco_extended_prefix}
\end{table*}

\FloatBarrier
\section{Sample hands across experimental conditions} \label{appx:samples}

In this section we present representative Dixit hands with the corresponding captions produced by the specialized speaker (after training) in our four main conditions, along with the corresponding distractors, listener match scores and rewards. We chose the speaker seed which achieved the highest overall score in that condition (though results were generally comparable across seeds), and to avoid cherry-picking examples we show the first four evaluation hands (from our randomly ordered evaluation) that contained a distinct target object (for example, in the crop condition, the first four that had a distinct object category in the bottom left). 

\begin{figure*}[p]
\centering
\includegraphics[width=0.62\textwidth]{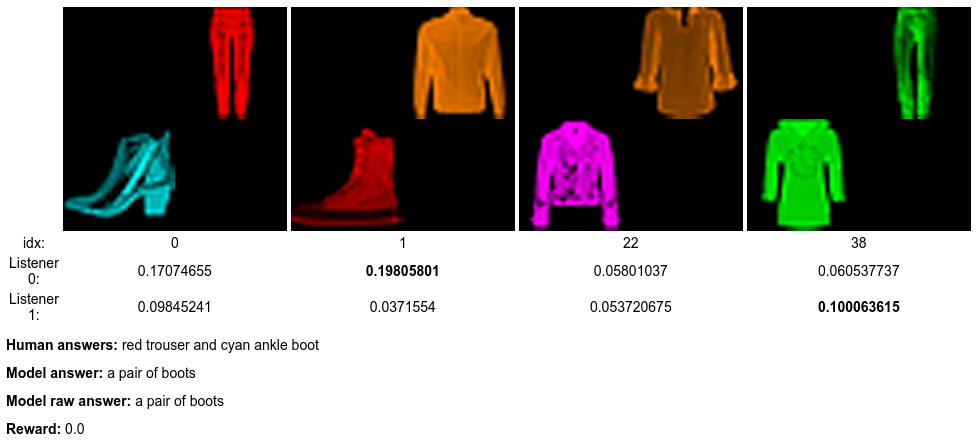}
\includegraphics[width=0.62\textwidth]{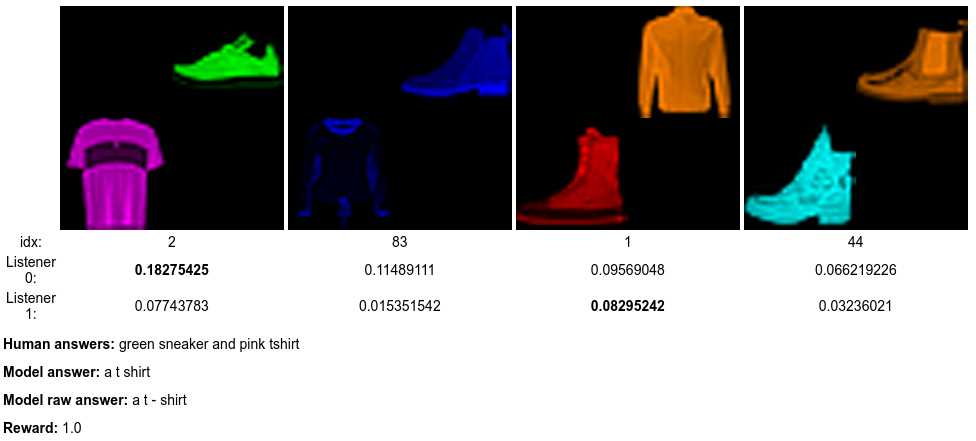}
\includegraphics[width=0.62\textwidth]{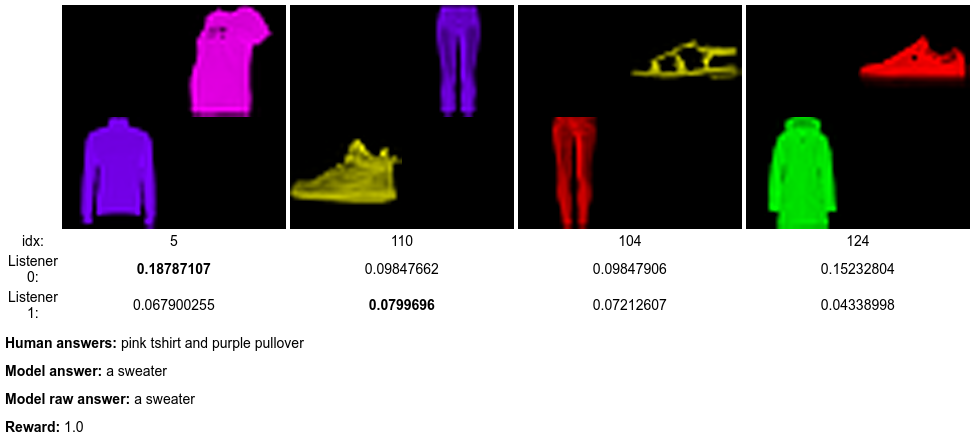}
\includegraphics[width=0.62\textwidth]{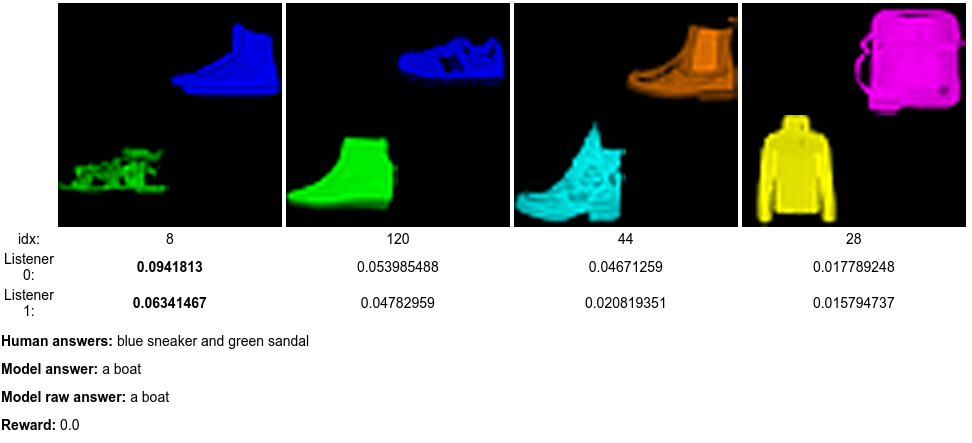}
\includegraphics[width=0.62\textwidth]{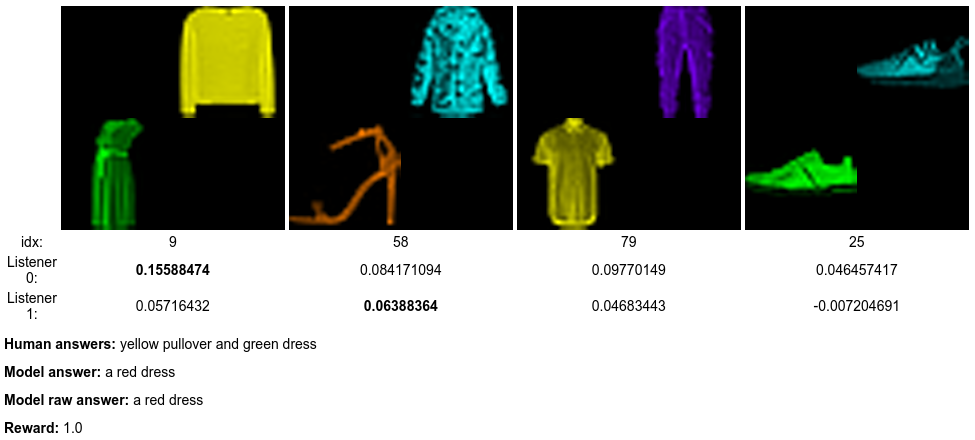}
\caption{Speaker samples and listener results on tcFMNIST, after the speaker has specialized to the second listener having the crop transformation. The language generally specifies the bottom left object, with a grounding failure in the second-to-last case (or perhaps a reference to boat sandals?).} \label{fig:samples_crop}
\end{figure*}

\begin{figure*}[p]
\centering
\includegraphics[width=0.62\textwidth]{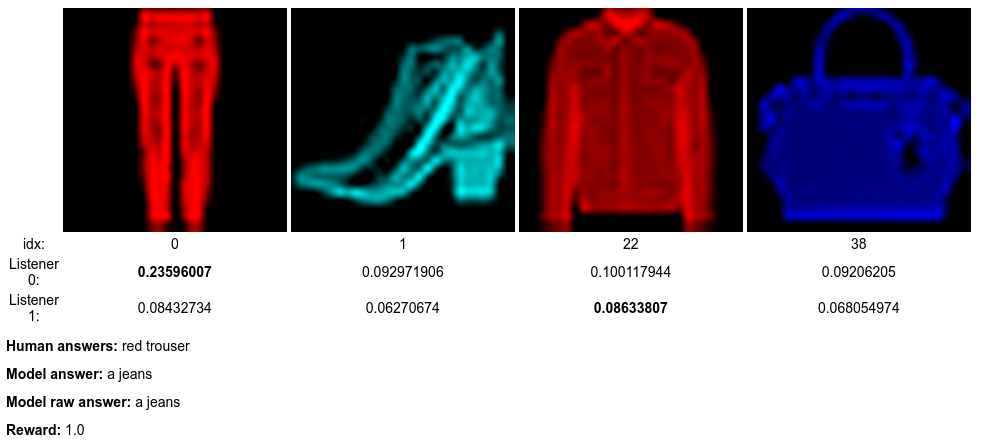}
\includegraphics[width=0.62\textwidth]{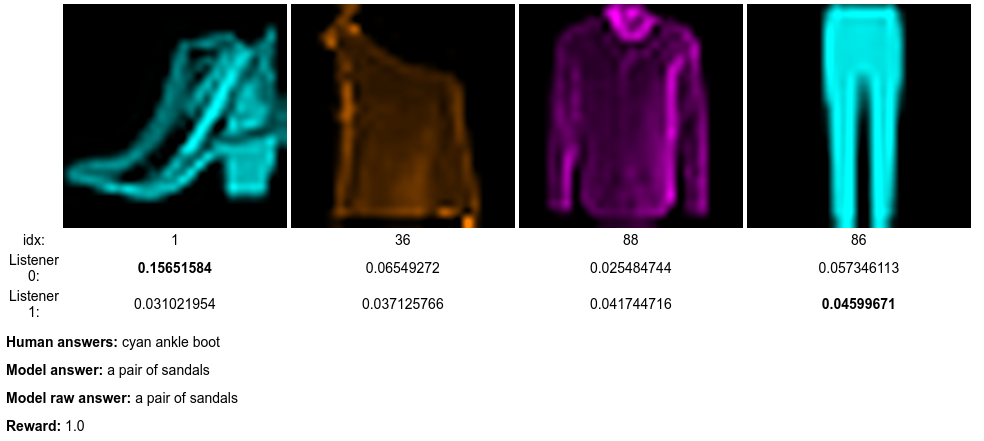}
\includegraphics[width=0.62\textwidth]{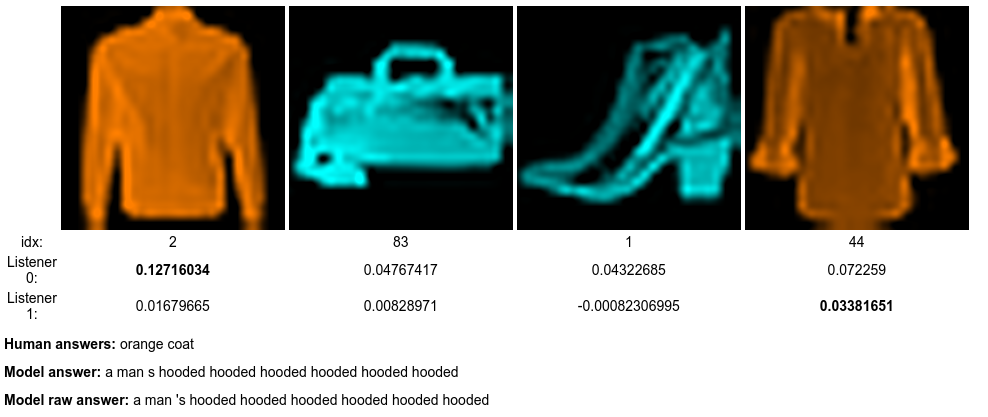}
\includegraphics[width=0.62\textwidth]{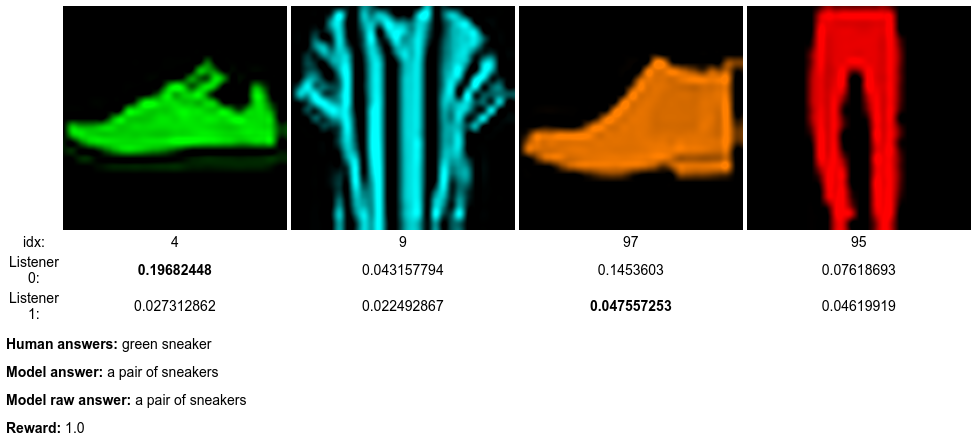}
\includegraphics[width=0.62\textwidth]{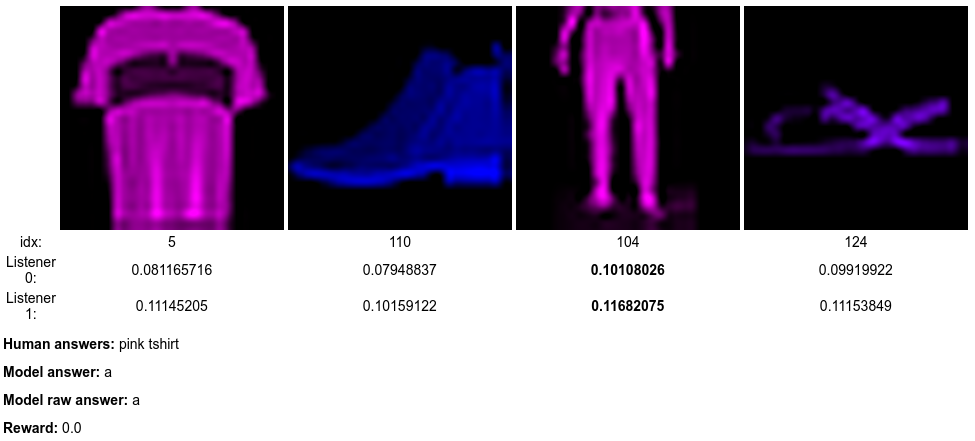}
\caption{Speaker samples and listener results on cFMNIST, after the speaker has specialized to the second listener having the blur transformation. The speaker generally ignores colors and names the object as intended, but with moderate language degradation in some cases.} \label{fig:samples_blur}
\end{figure*}

\begin{figure*}[p]
\centering
\includegraphics[width=0.62\textwidth]{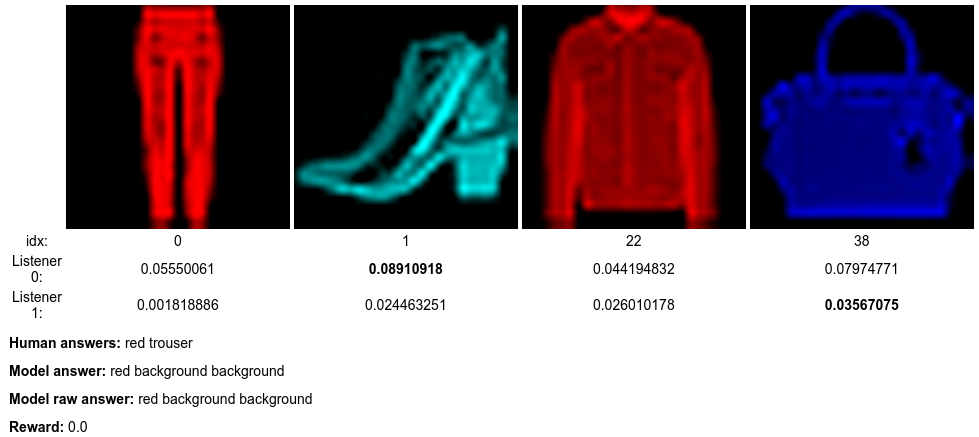}
\includegraphics[width=0.62\textwidth]{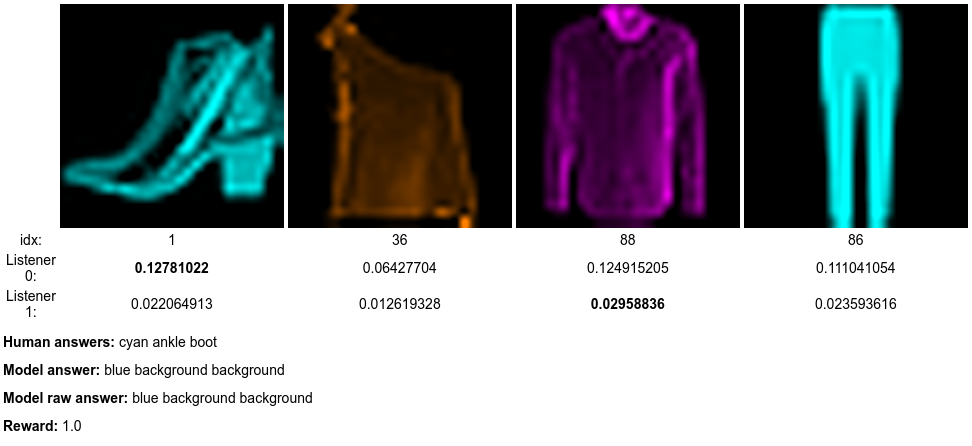}
\includegraphics[width=0.62\textwidth]{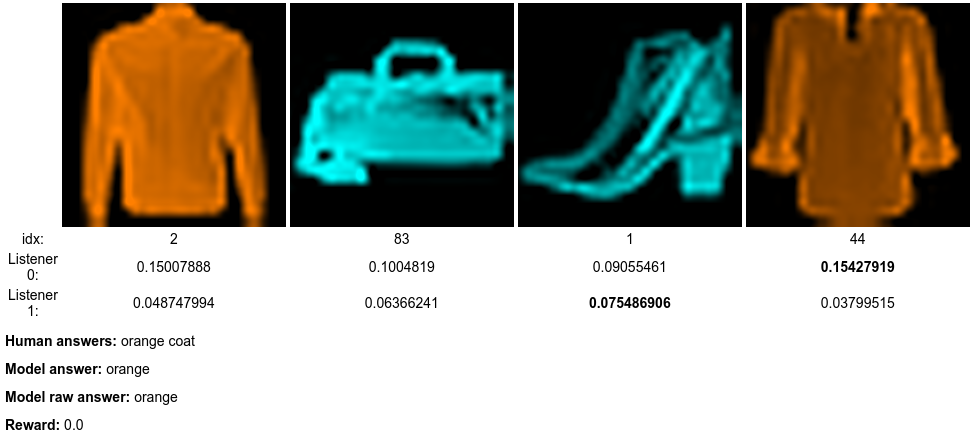}
\includegraphics[width=0.62\textwidth]{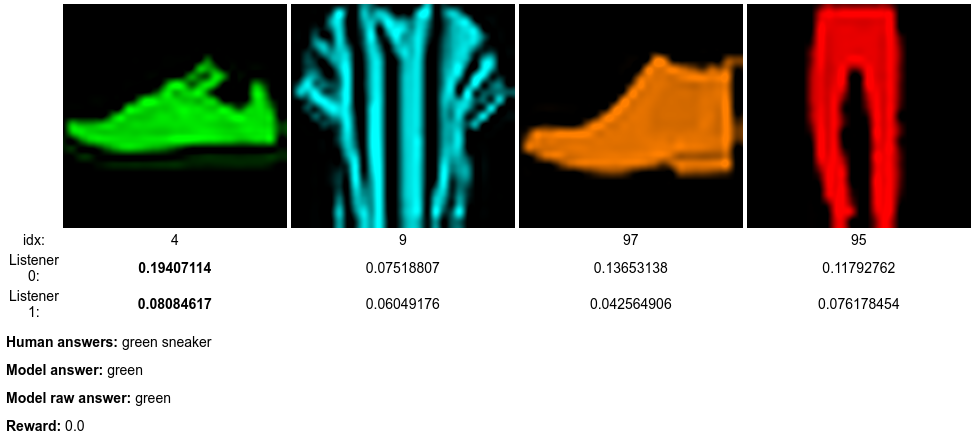}
\includegraphics[width=0.62\textwidth]{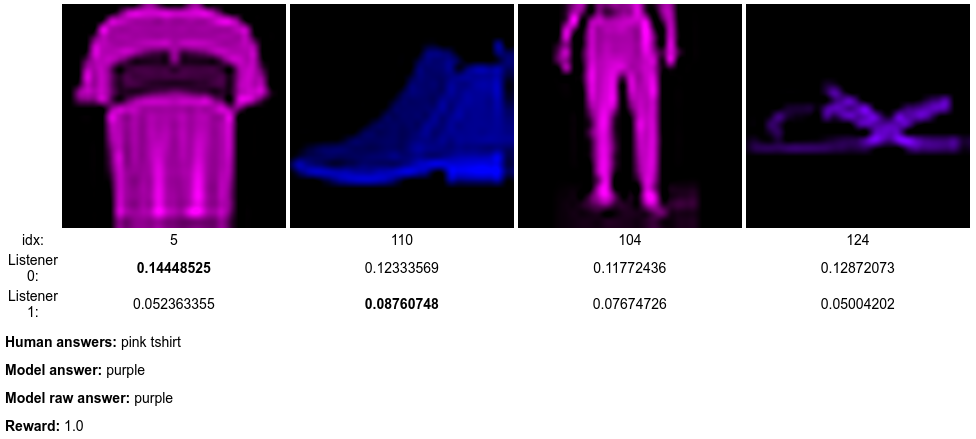}
\caption{Speaker samples and listener results on cFMNIST, after the speaker has specialized to the second listener having the grayscale transformation. The speaker consistently names the correct color, though it occasionally also repeats ``background''.} \label{fig:samples_gray}
\end{figure*}

\begin{figure*}[p]
\centering
\includegraphics[width=0.62\textwidth]{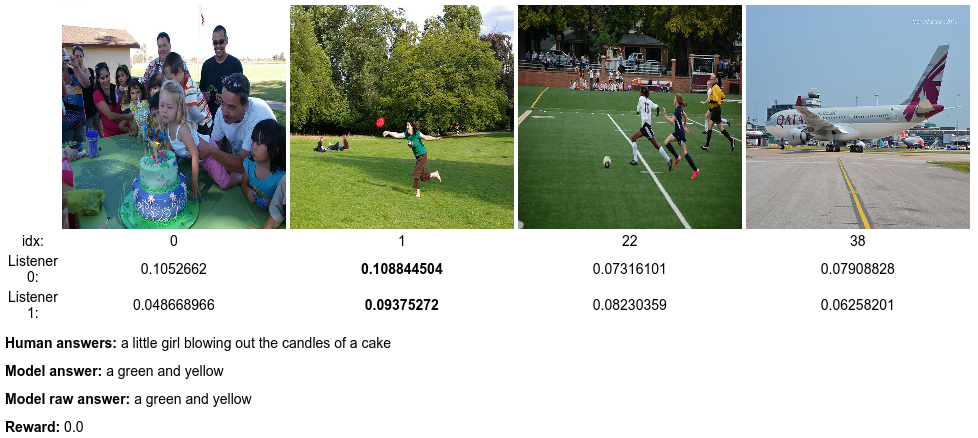}
\includegraphics[width=0.62\textwidth]{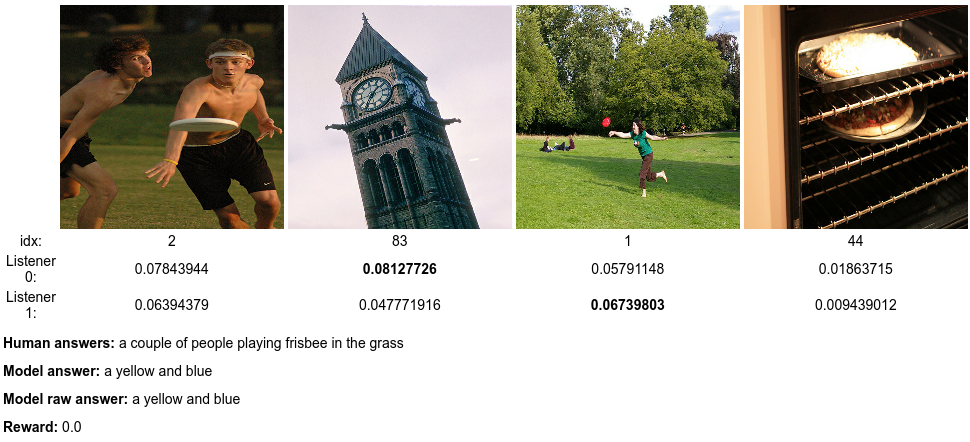}
\includegraphics[width=0.62\textwidth]{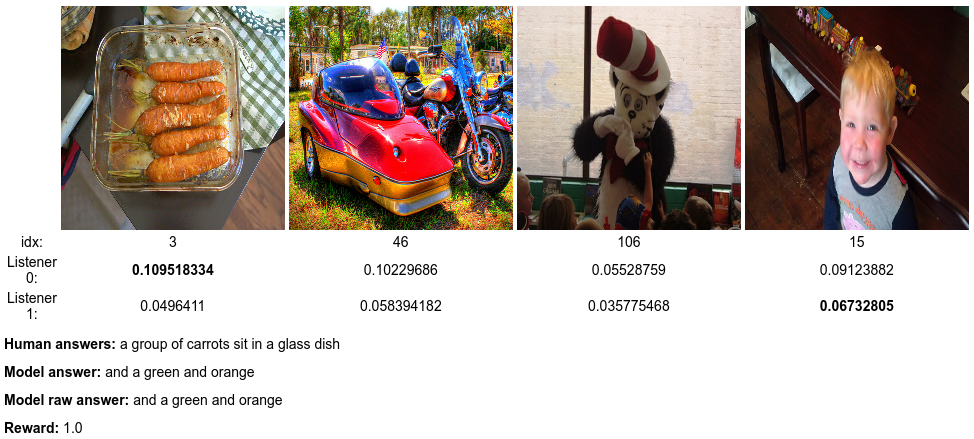}
\includegraphics[width=0.62\textwidth]{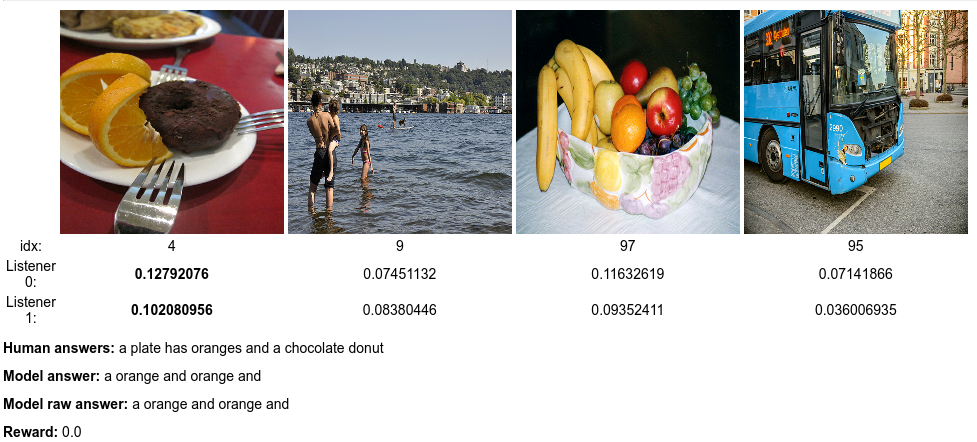}
\includegraphics[width=0.62\textwidth]{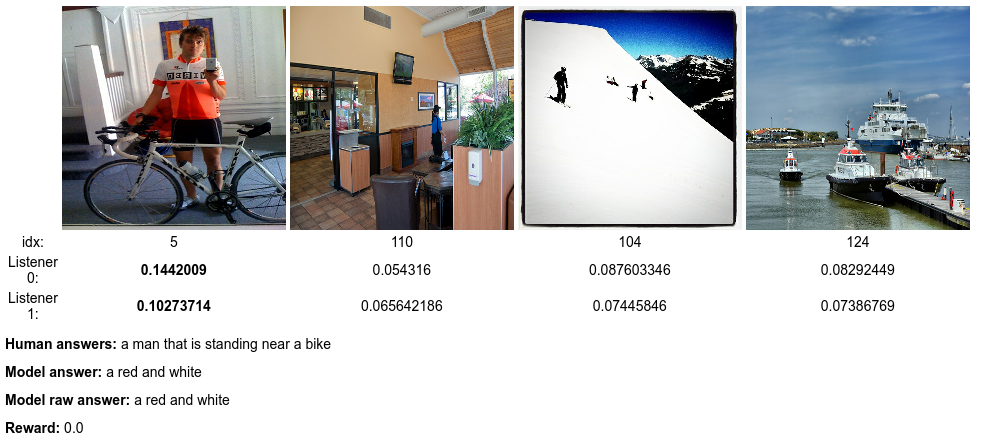}
\caption{Speaker samples and listener results on COCO, after the speaker has specialized (on COCO) to the second listener having the grayscale transformation. The speaker generally names one or two plausible colors for the images; but these are less discriminative in COCO than in the above results. There is also some minor language degradation in some cases (e.g. ``and" at the end of a caption).} \label{fig:samples_coco}
\end{figure*}

We present samples for the different transformation conditions in Figures \ref{fig:samples_crop}-\ref{fig:samples_coco}. Overall, the speaker adapts to the target difference between the listeners, and exhibits relatively mild language drift---the utterances generally stay grounded and human-interpretable, but in some cases exhibit some repetition or odd grammar. Some potential methods for further reducing these issues are noted above.

\textbf{Below-chance performance for the grayscale listener:} In Figure \ref{fig:resfig1}c, the grayscale listener achieves below-chance performance. We note from the samples (Figure \ref{fig:samples_gray}) that the speaker sometimes uses the word ``background'' in addition to using a color. This noun is not informative (so does not notably hurt the listener that can see color), but we speculate that it may throw off the second listener as some images have more background than others (e.g. a grayscale image of a high heel has more black than a grayscale image of a jacket). Note that the speaker does not see the distractors, so it can't purposefully consider the distractors to exploit this effect. This effect appears to be an artifact of the particular dataset we used, but does illustrate another intriguing way the speaker can exploit differences between listeners.

\end{document}